\documentclass[journal, twoside]{IEEEtran}
\usepackage{booktabs}
\usepackage{amsmath}
\usepackage{amssymb}
\usepackage{graphicx}
\usepackage{algorithm}
\usepackage{algorithmicx}
\usepackage{multirow}
\usepackage[numbers,sort&compress]{natbib} % Add lash between multiple refs. use '[1-3]' rather than '[1], [2], [3]'
\usepackage[noend]{algpseudocode}
\usepackage{hyperref} % Needs to be last package to avoid unkonwn exceptions.

\begin{document}

\title{3D Shape Knowledge Graph for Cross-domain 3D Shape Retrieval}

\author{Rihao Chang,~
        Yongtao Ma,~
        Tong Hao$^{\ast}$,~
        Weizhi Nie

\thanks{Yongtao Ma and Rihao Chang are with the School of Microelectronics, Tianjin University, Tianjin 300072, China. 
 (email: mayongtao@tju.edu.cn, changrihao@tju.edu.cn)}
\thanks{Weizhi Nie is with the School of Electrical and Information Engineering, Tianjin University, Tianjin 300072, China. (e-mail: weizhinie@tju.edu.cn)}
\thanks{*Corresponding author: Tong Hao is with the School of Life Sciences, Tianjin Normal University, Tianjin 300387, China.(e-mail: joyht2001@163.com)}
}

\maketitle

\begin{abstract}
  The surge in 3D modeling has led to a pronounced research emphasis on the field of 3D shape retrieval. Numerous contemporary approaches have been put forth to tackle this intricate challenge. Nevertheless, effectively addressing the intricacies of cross-modal 3D shape retrieval remains a formidable undertaking, owing to inherent modality-based disparities. This study presents an innovative notion—termed "geometric words"—which functions as elemental constituents for representing entities through combinations. To establish the knowledge graph, we employ geometric words as nodes, connecting them via shape categories and geometry attributes. Subsequently, we devise a unique graph embedding method for knowledge acquisition. Finally, an effective similarity measure is introduced for retrieval purposes. Importantly, each 3D or 2D entity can anchor its geometric terms within the knowledge graph, thereby serving as a link between cross-domain data. As a result, our approach facilitates multiple cross-domain 3D shape retrieval tasks. We evaluate the proposed method's performance on the ModelNet40 and ShapeNetCore55 datasets, encompassing scenarios related to 3D shape retrieval and cross-domain retrieval. Furthermore, we employ the established cross-modal dataset (MI3DOR) to assess cross-modal 3D shape retrieval. The resulting experimental outcomes, in conjunction with comparisons against state-of-the-art techniques, clearly highlight the superiority of our approach.
\end{abstract}


\IEEEpeerreviewmaketitle

%% main text
\section{Introduction}
The advancement of digitalization methods and computer vision has led to the widespread utilization of 3D shapes in various domains such as computer-aided design, medical diagnostics, bioinformatics, 3D printing, medical imaging, and digital entertainment. In recent years, there has been a desire for quick generation and simple access to vast quantities of 3D shapes, particularly for applications in virtual and augmented reality. It is reasonable to utilize some references to obtain similar 3D shapes and accelerate secondary development. These references can be 3D shapes~\cite{9127813,9055070}, 2D images~\cite{su2015multi}, sketch images~\cite{DBLP:journals/pr/LeiZZGML19}, and text information~\cite{DBLP:conf/jcdl/GoldfederA08}. Numerous methods have been put out in recent years to deal with this issue.

\begin{figure*}
    \centering
    \includegraphics[width=1\linewidth]{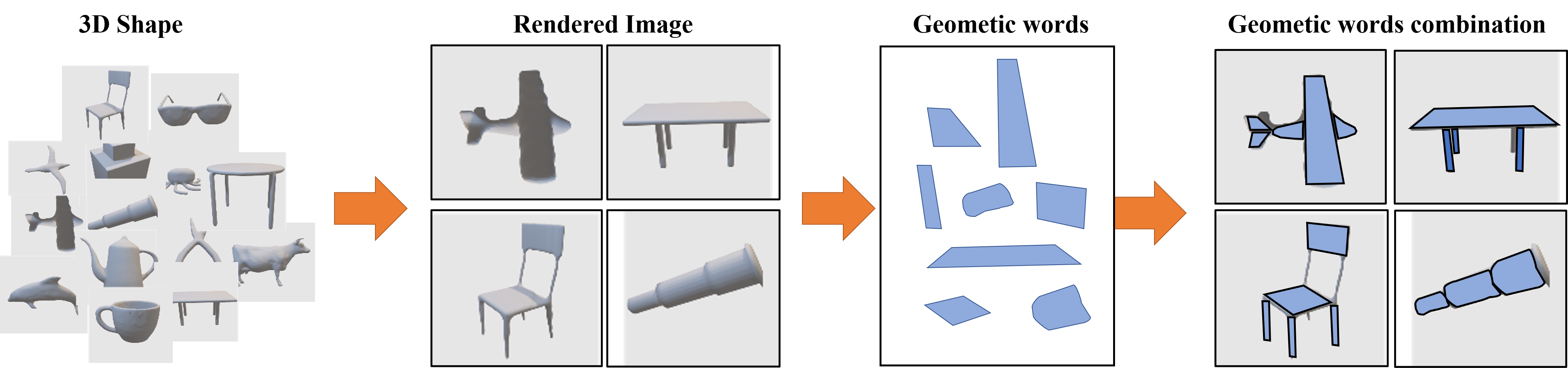}
    \caption{The schematic diagram for inspiration. Each 3D shape can be represented by a set of 2D views. The 2D image can also be split into a set of part shapes. Finally, these part shapes can be mapped into a standard geometric shape. This means that any 3D shape or a 2D image can be represented by finite geometric shapes if we have a complete dictionary of geometric shapes. Based on this assumption, we can handle the traditional 3D shape retrieval problem, 2D image-based 3D shape retrieval problem and cross-domain 3D shape retrieval problem according to the geometric dictionary.}
    \label{demo}
\end{figure*}

The MVCNN\cite{su2015multi} extracts a series of rendered views on 2D pictures and combines information from several perspectives of a 3D shape into a single, compact shape descriptor. PointNet++\cite{qi2017pointnet} iteratively implements the representations in a hierarchical neural network using point cloud representations for the input data to create a representation of three-dimensional forms. A multiloop-view convolutional neural network architecture for 3D shape retrieval was suggested by Gao et al.\cite{jiang2019mlvcnn}. It may be seen of as a modified MVCNN that takes into account the natural hierarchical links between views. Some researchers, however, concentrate more on the multimodal information to effectively represent the 3D shapes. You et al.\cite{you2018pvnet} suggested a combined convolutional network that successfully integrates point cloud and multiview images into an end-to-end neural network. These conventional methods concentrate on the descriptive layout of 3D shapes and use them to find related 3D shapes. However, consumers may now quickly obtain photographs thanks to the advancement of computer vision and smartphones. Some academics concentrate on the issue of retrieving cross-modal 3D shapes from 2D photos. Dai et al.\cite{Dai2018Deep} recommended a unique deep correlated holistic metric learning (DCHML) strategy to lessen the distinction between sketch images and 3D shapes. DCHML simultaneously trains modality-specific networks, mapping input data into an integrated feature space. Joint distribution adaptation (JDA), another transfer learning strategy, was put out by Long et al.\cite{long2014transfer}. It jointly adjusts the conditional distribution and the marginal distribution as part of a principled dimensionality reduction strategy to provide the effectiveness and robustness to significant distributional variations during feature learning. A coherent structure that minimizes the transition across domains statistically and geometrically has been proposed by Zhang et al.\cite{Zhang2017Joint}. In order to simultaneously decrease the geometric shift and the propagation shift, two combined projections are trained that project the information from the source domain and the target domain into low-dimensional subspaces. All of these techniques, however, concentrate on cross-modal feature learning and global structural information descriptor creation. Both of these depend on parameter learning, extensive training datasets, and model design. All of these techniques have trouble retrieving several cross-domain 3D models at once. These techniques cannot thus be seen as union methods.

\subsection{Motivation}
Traditional 3D shape retrieval techniques suffer from three distinct issues:1) They heavily rely on extensive 3D shape datasets to train network parameters. 2) They are ill-suited for addressing cross-domain and cross-modal retrieval challenges, where the efficacy of one modality diminishes when applied to different datasets. 3) None of these methods can effectively handle multi-condition cross-domain 3D model retrieval.
This paper introduces a novel concept to bridge these gaps. By identifying an intermediary variable capable of representing both 2D images and 3D shape information, we can establish connections between diverse domains and modalities. This intermediary variable serves as a representation of shape information and facilitates guided feature learning. Thus, the task at hand is to determine how to identify and utilize this intermediary variable.

This paper introduces a novel concept to bridge these gaps. By identifying an intermediary variable capable of representing both 2D images and 3D shape information, we can establish connections between diverse domains and modalities. This intermediary variable serves as a representation of shape information and facilitates guided feature learning. Thus, the task at hand is to determine how to identify and utilize this intermediary variable.

In this paper, a novel idea is presented. If we can find an intermediate variable, which can be used to represent the 2D image and the 3D shape information, we will be able to bridge the gap between different domains and different modalities. Naturally, this variable can be used to represent the shape information and help to guide feature learning. Thus, we need to think about how to find this intermediate variable. 

In Fig.\ref{demo}, a single 3D shape finds representation through a set of rendered images. Additionally, these rendered images can be deconstructed into a set containing geometric information. For example, a "cup" can be reduced to a cylinder, and a "table" can be decomposed into a square plane and a cylinder. When we treat a 3D shape as a document, its rendered images can be likened to sentences, while geometric information constitutes the "geometric words." These "geometric words" represent elemental shape components, allowing any shape to be expressed through a limitless combination of such words. Importantly, the term "geometric word" surpasses the conventional interpretation of geometric primitives, which encompasses standard shapes like cylinders, squares, and triangles. Instead, the concept of a "geometric word" boasts a broader scope, applicable to any shape. This perspective liberates us from relying on an extensive array of 3D shapes for training, as we can sufficiently describe shapes by generating an ample assortment of geometric words. This technique has the potential to address cross-domain retrieval concerns, as each 3D shape can be delineated by an infinite array of geometric words. Furthermore, geometric information can be extracted from 2D images, bridging the gap between 2D images and 3D shapes and thus facilitating cross-modal 3D shape retrieval. Nonetheless, we are confronted with two pivotal challenges: 1) Defining and identifying the comprehensive set of "geometric words" serving as a holistic shape representation. 2) Devising methods for shape retrieval and cross-domain shape retrieval predicated upon these "geometric words."

This paper, grounded in the notion of "geometric words," proposes an innovative 3D shape knowledge graph and graph embedding technique to address the complexities of cross-domain 3D shape retrieval. We begin by leveraging OpenGL to construct a toolbox\footnote{Removed for anonymized peer review} for generating a set of rendered 3D shape images. Subsequently, we deploy an image segmentation approach to extract part shapes. The unsupervised technique of K-means is then employed to identify geometric words based on these part shapes. At this juncture, we have acquired a collection of rendered images and associated part shapes. The subsequent steps encompass the construction of a 3D shape knowledge graph that encapsulates the relationships among 3D shapes, rendered images, part shapes, and geometric words. A unique graph embedding strategy is subsequently introduced to facilitate the learning of embeddings for 3D shapes, rendered images, shape parts, and geometric words, incorporating their inherent structural attributes. Finally, an effective similarity measurement methodology is proposed to address 3D shape retrieval. Notably, our approach adeptly addresses cross-domain and cross-modal 3D shape retrieval in one unified framework, all the while requiring minimal data to establish the 3D shape knowledge network.

\subsection{Contributions}
The contributions of this paper are outlined as follows:
\begin{itemize}
\item We present a pioneering 3D shape knowledge graph that proficiently tackles the intricate challenges of cross-domain 3D shape retrieval. To the best of our knowledge, this study marks the inaugural application of the knowledge graph paradigm to the realm of 3D shape retrieval, uniquely addressing multiple cross-domain aspects.
\item A novel graph embedding strategy is proposed for representing entities within the 3D shape knowledge graph. This strategy effectively addresses representation learning under both supervised and unsupervised conditions.
\item A similarity metric between query shapes/images and target 3D shapes, rooted in knowledge graph embeddings, is introduced. This metric adeptly incorporates geometric structural and categorical information of 3D shapes, resulting in enhanced retrieval precision.
\end{itemize}

The subsequent sections delineate the remaining content of this document. Section 2 expounds upon related work, while Section 3 introduces our proposed solution. Section 4 showcases experimental outcomes, elucidating the effectiveness of our approach. A comprehensive review of the results is also furnished in this section, employing our approach to confront a range of 3D shape retrieval challenges and demonstrate its efficacy. Lastly, Section 5 delves into potential avenues for future research.

\section{Related Work}
In recent years, there has been a surge in the development of diverse 3D shape recognition methods. This section introduces classic 3D shape retrieval methods and recent advancements in cross-domain 3D shape retrieval.

\subsection{3D Shape Retrieval}
The advancement of 3D data processing technology has led to the application of 3D shape retrieval across various domains. To convolve a 3D shape akin to any other tensor \cite{Wu20153D,qi2017pointnet++}, several studies have concentrated on voxelized shapes. These techniques encounter resolution constraints due to data sparseness and the computational expense associated with 3D convolution.

For the purpose of 3D object identification and retrieval, Garro et al. \cite{DBLP:journals/pami/GarroG16} introduced tree-based shape representations grounded in specific graph kernels and the scale space of the autodiffusion function (ADF), enabling the incorporation of texture and other structural information. Xie et al. \cite{DBLP:journals/pami/XieDZWF17}, by estimating multiscale shape distributions and applying Fisher discriminant criteria to neurons, proposed a discriminative deep autoencoder for learning deformation-invariant shape information. PointNet \cite{qi2017pointnet} pioneered a technique for direct processing of point clouds using deep neural networks, albeit disregarding local features. These methodologies exhibit commendable performance on certain publicly available 3D shape datasets. Nevertheless, these approaches primarily rely on mathematical models to derive effective structural representations, neglecting geometric information of practical significance.

In the context of addressing classification and retrieval challenges, Wang et al. \cite{wang2019dynamic} introduced the EdgeConv module, suitable for point cloud tasks integrated with CNNs. Concurrently, other strategies tackle 3D representation using multiview data. The lighting field descriptor (LFD) \cite{LFD} serves as an initial viewpoint-based 3D representation, comparing the corresponding 2D properties of two view sets to ascertain the similarity between 3D objects. Similarly, GIFT \cite{GIFT} calculates the Hausdorff distance between their corresponding view sets. Traditional methods of 3D shape representation can be regarded as variants of LFD and GIFT. Su et al. \cite{su2015multi} recently introduced a multi-view convolutional neural network (MVCNN), which generates numerous 2D projection features through CNN-based learning in a trainable end-to-end manner. Sfikas \cite{panoramann} proposed a technique for capturing panorama view features, aiming to ensure 3D shape continuity and minimize data preprocessing by creating an augmented image representation.

In recent years, graph structures have gained prominence in handling the representation of 3D models. Utilizing a graph neural network, Wang et al. \cite{wang2018LocalSpecGCN} introduced local spectral graph convolution to jointly consider point information and information from neighboring points. Te et al. \cite{Te2018RGCNN} employed spectral graph theory to develop a regularized graph convolutional neural network. This network maps the point cloud onto a graph structure and performs calculations on this structure using Chebyshev polynomial approximation.

To address classification and retrieval challenges, Wang et al. \cite{wang2019dynamic} proposed EdgeConv, an approach suitable for point cloud tasks within a CNN framework. Shi et al. \cite{Shi2020Point-GNN} devised Point-GNN to mitigate translation variance and introduced a process of box merging and scoring to accurately aggregate detections from various vertices. Zhang et al. \cite{Zhang2021SGG} introduced an edge-oriented graph convolutional network that leverages multidimensional edge information for relationship modeling and the study of interactions between nodes and edges. These methodologies emphasize the utilization of structural information to enhance the performance of 3D shape features. However, these graph-based approaches tend to overlook cross-modal relation information and local details of 3D shapes.

Both of these methods have developed corresponding networks for learning 3D form representations using prevalent deep learning techniques. These approaches rely extensively on abundant training data, rather than geometric structural data. Adapting these methodologies to address the challenge of cross-domain 3D shape retrieval presents a formidable task.

\subsection{Cross-domain 3D Shape Retrieval}
Cross-domain refers to the exploration that enable effective knowledge transfer \cite{zhang2022tn,li2023object} and generalization between different, often unrelated, modalities or data sources \cite{li2022video}, facilitating improved performance and insights across various tasks.
Image-based retrieval has emerged as a contemporary technique with the potential to become a formidable competitor \cite{he2022category,li2018deep,li2015weakly}. Mu et al. introduced a novel paradigm for image-based 3D shape recovery \cite{mu2018image-based}. The approach initially represents the image as a Euclidean point, transforming all displayed views of a 3D shape into Symmetric Positive Definite (SPD) matrices, effectively representing them as points on a Riemannian manifold. The recovery of image-based 3D shape is then simplified into a process of learning Riemannian metrics from Euclidean metrics.
Li et al. constructed an embedding space using a 3D shape similarity measure \cite{li2015joint}. They further employed a Convolutional Neural Network (CNN) to enhance the purity of images by eliminating distracting elements. This joint embedding strategy enables cross-view image retrieval, image-based shape retrieval, and shape-based image retrieval. Despite the limited availability of relevant works in the image retrieval community, these techniques have been established and serve as important references.
Incorporating a Generative Adversarial Network (GAN) into the process, random noise drawn from a fixed distribution can be harnessed to generate coherent images while handling function transformations. The forthcoming sections provide detailed explanations of these methodologies. 
It is noteworthy to consider the definition of SeqViews2Seqlabels \cite{han2019seqviews2seqlabels:} during the feature extraction phase.To capture the global characteristics of 3D shapes, the SeqViews2Seqlabels model is introduced. This model maintains spatial and content knowledge across sequence views through aggregated sequences. Simultaneously, by adjusting the weight of specific views, the discriminative capacity of the SeqViews2Seqlabels model is enhanced. Upon closer examination, these approaches collectively constitute cross-domain feature learning. Their primary objective is to facilitate feature learning from multi-modal data within an embedding space. Notably, these approaches tend to overlook geometric information of practical significance.

\begin{figure*}[ht]
	\centering
	\includegraphics[width=1\linewidth]{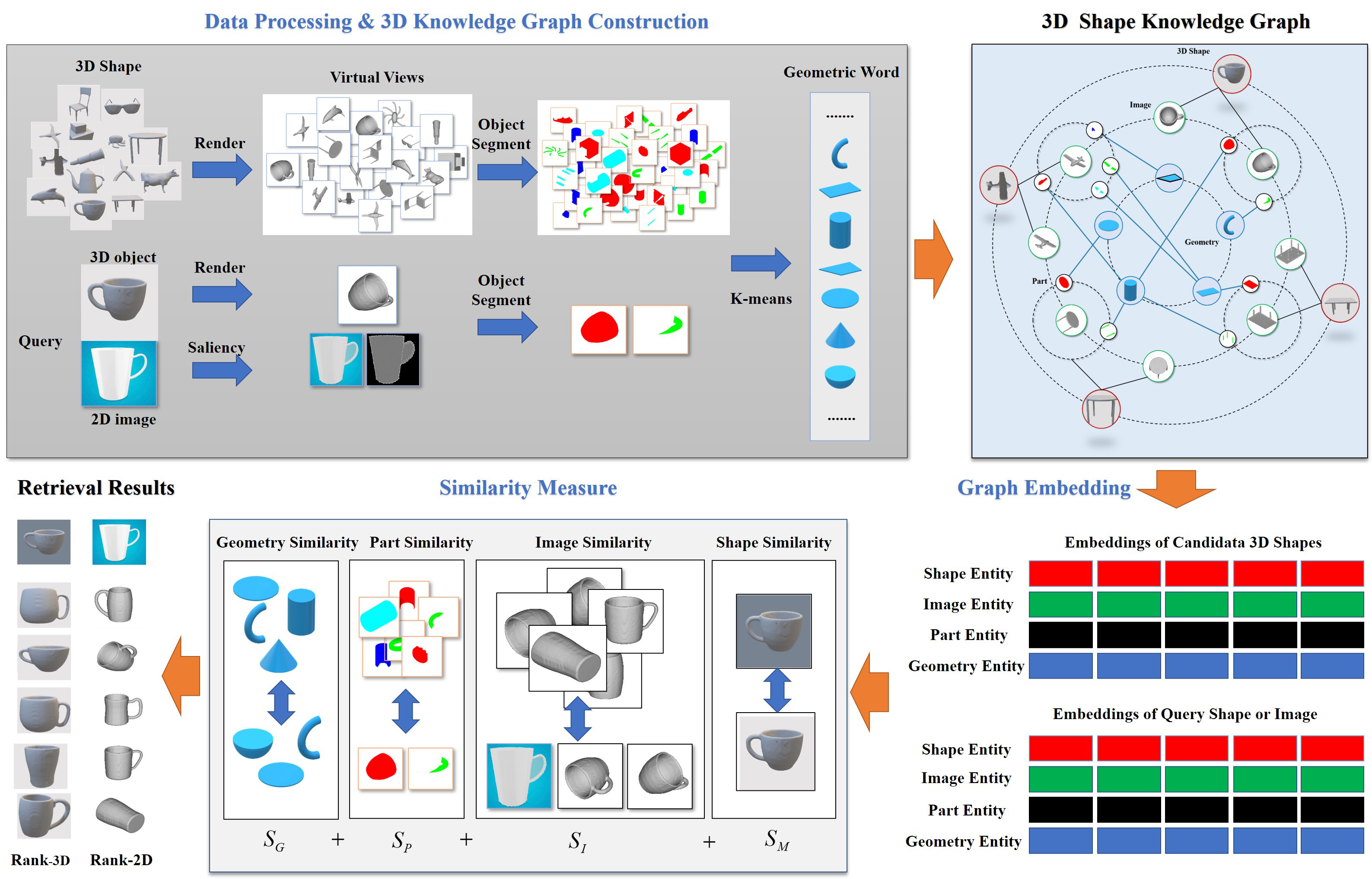}
    \caption{The framework of our approach, which includes three parts: knowledge graph construction (entity and edge extraction), graph embedding and similarity measure. K-means is used to generate geometric descriptions. The embedding strategy embeds the graph entities into vector space for retrieval.}
	\label{graph23}
\end{figure*}

\subsection{3D Model based on Component Theory}
In this paper, we introduce a "geometric word" theory inspired by previous research \cite{nie20203d}, which serves as a versatile framework for representing shape information. Conceptually, a "geometric word" can be conceived as a fundamental constituent of an object, akin to the notion of components introduced in prior work on component theory.
Liu et al. \cite{liu2018physical} presented an innovative formulation that encompasses the learning of physical primitives to expound both an object's visual attributes and its behavior within physical events. Their approach offers a compelling strategy for tackling segmentation challenges, particularly in the context of tool behaviors, while remaining adaptable to more traditional models.
Katageri et al. \cite{katageri2021pointdccnet} proposed the Point Decomposition Network (PointDCCNet) tailored for 3D object classification. This method relies critically on the performance of its decomposition module.
Our methods are further inspired by certain multi-view 3D model segmentation strategies, such as those outlined in \cite{ahmed2020rgb,zhuang2021semantic,wang2020unequal}. These approaches transform 3D point clouds into 2D images, effectively translating the 3D analysis into a 2D problem addressable through CNN-based solutions.
Feng et al. \cite{feng2018gvcnn} introduced the Group-view Convolutional Neural Network (GVCNN) as a means of hierarchical correlation modeling aimed at delivering discriminative 3D shape descriptions. By providing multiview information through a grouping mechanism, each group can be interpreted as a distinctive component. This design effectively eliminates redundant information, ultimately enhancing the final performance.
Mo et al. \cite{mo2019partnet} introduced an extensive 3D object dataset featuring intricate annotations and structured object parts. This dataset offers a valuable lens through which to comprehend 3D models and served as an inspiration for our formulation of the geometric word theory.
In recent years, a surge of research has delved into the realm of fine-grained and hierarchical shape segmentation. Yi et al. \cite{yi2017learning} leveraged noisy part decomposition derived from CAD model designs to learn consistent shape hierarchies. Furthermore, a recursive binary decomposition network \cite{yu2019partnet} was introduced to address shape hierarchical segmentation challenges in a recursive manner.

\section{Our Approach}
This section offers a comprehensive introduction to our approach, as illustrated in Fig. \ref{graph23}. The complete framework consists of three core steps:
\begin{itemize}
\item  Construction of the 3D Shape Knowledge Graph: The initial step involves generating multiple images from various viewpoints of 3D shapes. Subsequently, we employ an image segmentation technique \cite{Kalogerakis_2017_CVPR} to partition each rendered image into distinct shape parts. These segmented parts are then categorized into a collection of geometric words, allowing for the projection of each shape component into a corresponding geometric word. Additionally, we establish entities and edges to facilitate the construction of the 3D shape knowledge graph.
\item  Graph Embedding: Our approach introduces a graph embedding strategy tailored to the structure of our 3D shape knowledge graph. Notably, we define the category edge within the knowledge graph. This edge's manipulation enables our approach to effectively address both supervised and unsupervised problems.
\item  Similarity Measure: Drawing on the entity embeddings, we put forth an efficient similarity measurement method. This method encompasses various similarity measurement strategies, as depicted in Fig. \ref{graph23}. Further elaboration on these steps can be found in Subsection \ref{simeasure}.
\end{itemize}
The ensuing sections will provide a detailed breakdown of each of these steps, elucidating our approach's intricacies and contributions.

\subsection{Data Preprocessing}
Data processing plays a pivotal role in our approach, focusing on the crucial task of identifying visual geometric words to construct the 3D shape knowledge graph. This process encompasses three distinct steps: 1) Extraction of Rendered Views: Starting with a 3D model, we extract multiple rendered views. Each 3D model generates a corresponding set of images. 2) Image Segmentation: We apply an image segmentation technique to segment the rendered images. Each object is dissected into a collection of parts, with each part associated directly with its corresponding rendered image. 3) Part Classification for Geometric Words: Subsequently, we classify these parts to extract essential information. The objective here is to discern the geometric words, with each class representing a distinct geometric word. The central point of each part serves as the representative of its associated geometric word.

In the course of this process, our focus rests on 2D object segmentation techniques \cite{li2021ctnet}. Classic 3D shape decomposition methods or unsupervised techniques such as those in \cite{liu2018physical, burgess2019monet, nguyen2019hologan} do not apply in this context. The absence of an effective method for object segmentation led us to develop a new 3D shape segmentation dataset. To train the segmentation model, we employed the well-established FCN model \cite{Kalogerakis_2017_CVPR}.
The segmentation dataset, as depicted in Fig. \ref{fig:fig1}, along with its detailed specifications available in the supplemental files, has been made publicly accessible on GitHub\footnote{Removed for anonymized peer review}. We employed the classic FCN model \cite{Kalogerakis_2017_CVPR} for training, utilizing 4,940 samples for training, 1,900 for validation, and 760 for testing. Notably, Fig. \ref{fig:fig2} showcases select segmentation results.
During training, unlike traditional segmentation models, our approach doesn't require object-specific label information, relying solely on the ground truth of segmentation. This implies that the segmentation model trained using our dataset serves as a general segmentation tool applicable to diverse shapes. This process ultimately yields a collection of shape parts associated with each rendered or real image, as illustrated in Fig. \ref{fig:fig2}. Several well-regarded segmentation techniques were also employed for comparative analysis. The outcomes of these experiments are presented in Table \ref{table:table2}.
The retrained model exhibits superior performance. Through these processes, we attain sets of rendered images, shape parts, and geometric words, each with explicit correlations. This valuable information forms the foundation for constructing the 3D shape knowledge graph. Importantly, it's worth acknowledging that segmentation methods serve as preprocessing steps. Their performance significantly influences the eventual outcomes of retrieval and classification. While other segmentation methods could be employed, we have selected the most effective approach for our subsequent work.

\begin{figure}[ht]
    \centering
    \includegraphics[width=1\linewidth]{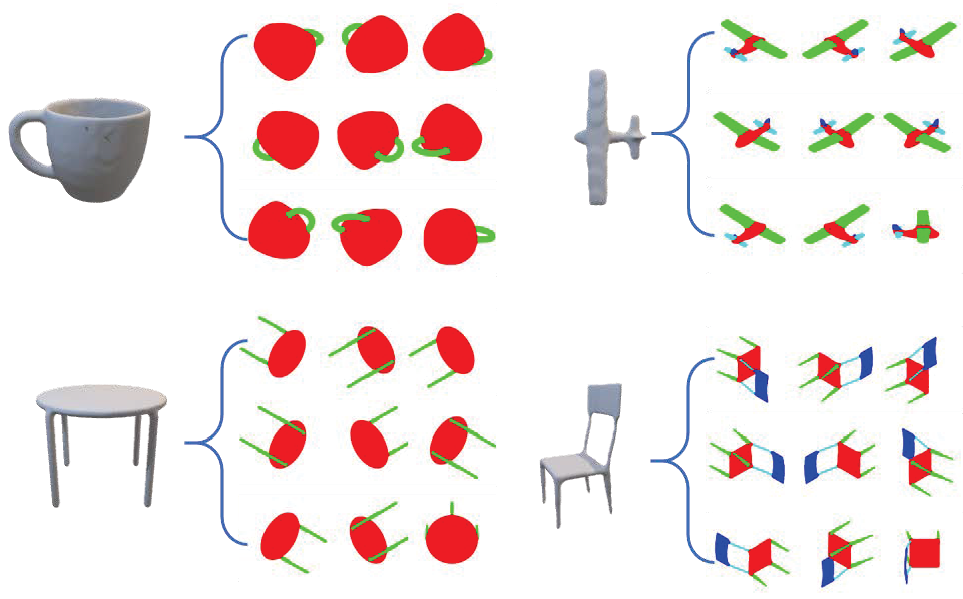}
    \caption{Some training examples in the object segmentation dataset}
    \label{fig:fig1}
\end{figure}

\begin{figure}[ht]
    \centering
    \includegraphics[width=1\linewidth]{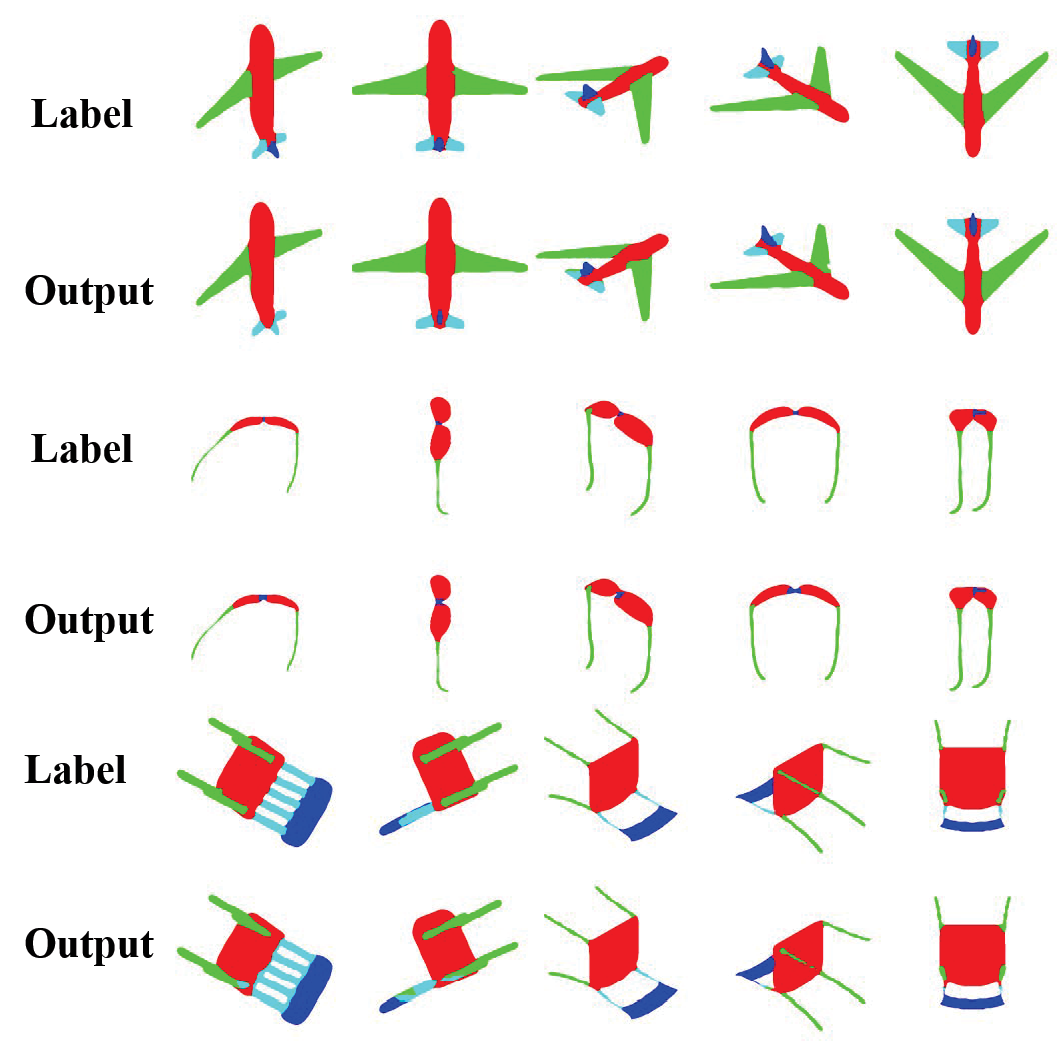}
    \caption{Some segmentation results based on the FCN}
    \label{fig:fig2}
\end{figure}

\begin{table}[ht]
  \centering
  \caption{Performance comparisons of the FCN with other classical methods on our dataset}
  \label{table:table2}
    \begin{tabular}{lcc}
      \toprule
      $Model$ &$Backbone$ &$MIoU$\\
      \midrule
      U-Net\cite{10.1007/978-3-319-24574-4_28}	&-	&80.68\\
      SegNet\cite{2017SegNet}	&VGG16	&82.84\\
      PSPNet\cite{2017Pyramid}	&ResNet-103	&86.49\\
      DeepLab v3\cite{2017Rethinking}	&ResNet-101	&90.42\\
      FCN	&VGG16	&\bf96.09\\
      \bottomrule
    \end{tabular}
\end{table}

\subsection{3D Shape Knowledge Graph Construction}
This section elucidates the meticulous process of constructing the 3D shape knowledge graph, which necessitates the definition of entities and their associated edge information. The sufficiency of edges or relations within the graph significantly impacts the representation of the 3D shape, as limited information hinders the effective learning of entities. Hence, the architectural structure of the knowledge graph is of paramount importance, and this process is detailed comprehensively below.

Entity Definitions in the 3D Shape Knowledge Graph:

1. Model Entity: This entity embodies the query 3D shape. We employ PointNet++ \cite{qi2017pointnet++} to extract feature vectors for each shape and subsequently integrate them into $G$. To maintain uniform dimensionality of shape descriptors, PCA \cite{wold1987principal} is applied for feature vector dimension reduction.

\begin{figure}[ht]
	\centering
	\includegraphics[width=1\linewidth]{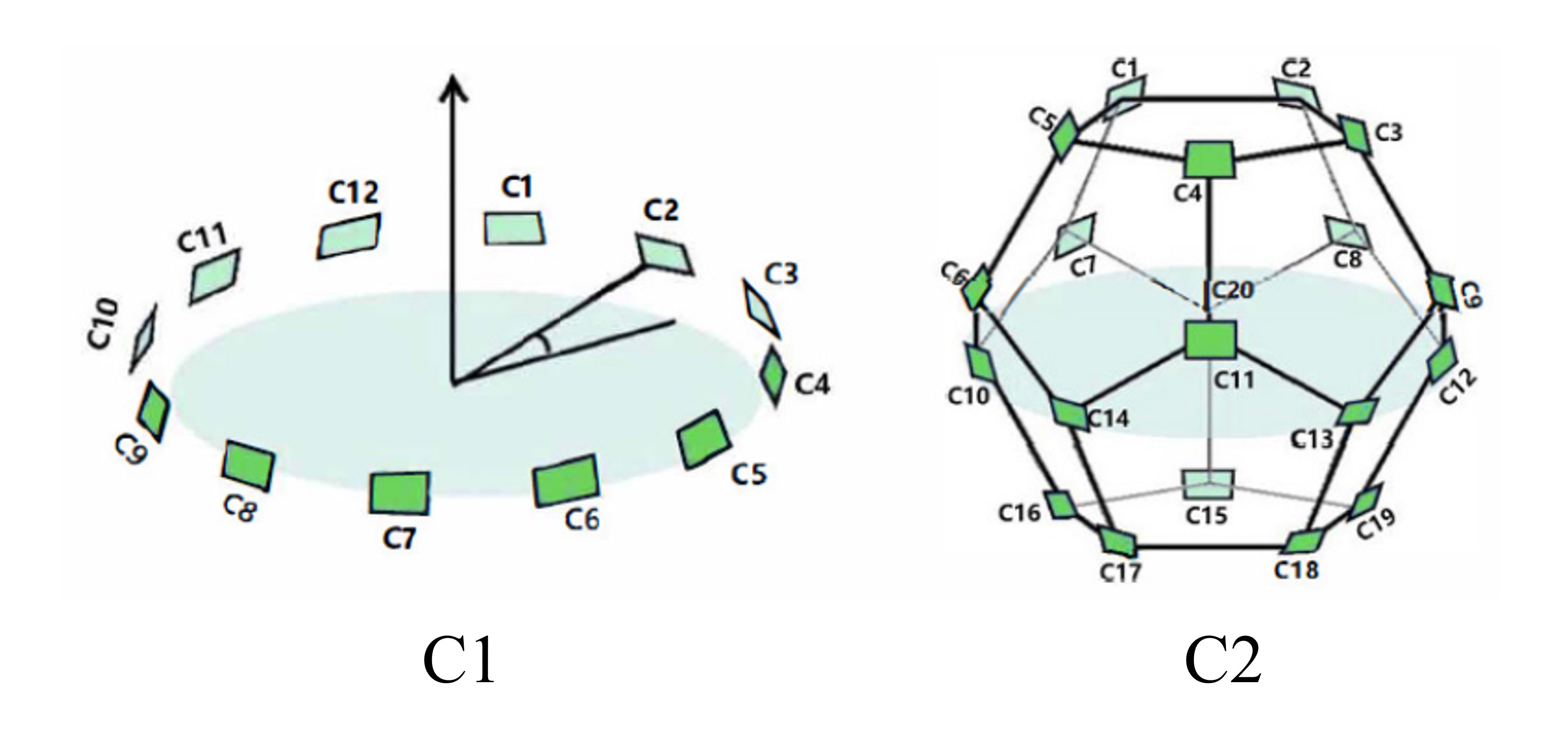}
	\caption{Illustration of classic viewpoint setups. Both of C1 and C2 center the 3D object and surround the object with cameras for comprehensive views.}
	\label{c1}
\end{figure}

2. Image Entity: This entity represents rendered images derived from the 3D shape. Furthermore, it can represent real images if the knowledge graph is used for cross-modal 3D shape retrieval. We utilize two methodologies for extracting rendered images, as delineated in Fig. \ref{c1}.
For Case C1, we employ NPC \cite{npca} to orient the 3D shape upright along a fixed axis (e.g., z-axis). Cameras are then positioned at a fixed angle $\theta$ around this axis, set at 30 degrees from the ground plane, and facing the shape's center. Different $\theta$ values generate varying views, producing $\{20; 16; 12; 10; 8; 6; 4; 2; 1\}$ views for each object.
In Case C2, a diverse perspective is adopted, where shapes are not continuously kept upright. Instead, multiple views are sampled from the 3D space. Specifically, we deploy 20 virtual cameras at the vertices of a dodecahedron shape situated around the object's center.

3. Part Entity: Based on rendered or real images, each image is segmented into parts using a model trained on our dataset \cite{Kalogerakis_2017_CVPR}. These segmented parts serve as part entities, representing the attributes of each shape in the knowledge graph. Real images require additional preprocessing due to complex backgrounds, where salient object detection \cite{wang2019salient} is applied to separate the object from the background.

4. Geometric Word Entity: These entities serve as pivotal elements in the 3D shape knowledge graph, bridging the gap between disparate domains or modalities. Geometric word entities are generated from part entities originating from different images, including rendered images and real images representing the same object. Similar part entities are categorized as geometric words. We employ a pretrained CNN \cite{he2016identity} to generate descriptors, followed by K-means clustering to determine geometric word labels, with their centers serving as descriptors.

Edge Definitions in the Knowledge Graph :

1. Lash Edge: This edge establishes connections between the 3D shape and its rendered images, rendered images and segmented parts, and real images and segmented parts. Such connections embody the geometric and visual attributes of the 3D shape within $G$.

2. Geometric Edge: This edge links part entities to their respective geometric word entities, reflecting the geometric structure of the 3D shape or real image.

3. Category Edge: Capturing inter-category relations, this edge represents a priori knowledge. Its presence allows our approach to function in a supervised manner. Removal of this edge renders our approach unsupervised.

In summary, the knowledge graph $G$ holistically captures the geometric structure and categorical information of 3D shapes. Each edge encapsulates distinct knowledge, e.g., "table, rectangle, lash" signifying the presence of a rectangle within a table. Notably, entities (3D shapes, 2D images, geometric words) lack explicit text labels. As depicted in Fig. \ref{demo}, the segmentation process indirectly illustrates the composition of the knowledge graph.

\subsection{Graph Embedding}
Up to this point, we have successfully constructed the 3D shape knowledge graph. To effectively address the challenges of cross-domain 3D shape retrieval, we must devise a strategy to generate embeddings for each node based on the structure of this knowledge graph. This section provides an in-depth explanation of our innovative embedding approach.

\subsubsection{Problem Definition}
The 3D shape knowledge graph captures the structural aspects of 3D shapes and directly reveals the correlation between visual information and 3D shapes. However, it does not fully address the weak correlation that exists in (Euclidean) space between a real image and a 3D shape. Consequently, employing a graph embedding technique to create dense embedding features for sparse data becomes a compelling avenue.

In this study, we introduce a novel graph convolutional network to generate node embeddings, thereby leveraging the structure of the 3D shape knowledge graph, particularly the geometric word, to enhance feature vector consistency between similar 3D shapes and to establish connections between images and 3D shapes.

We define the knowledge graph as an undirected weighted graph $G=(V,E)$, where $V$ denotes the nodes and $E$ represents the edges. Specifically, we divide the node set $V$ into five distinct parts: $V={V^M, V^I_v, V^I_r, V^P, V^G}$, where $V^M$ signifies the model entity set, $V^I_v$ corresponds to the rendered image entity set, $V^I_r$ pertains to the real-image entity set, $V^P$ encompasses the part shape entity set, and $V^G$ represents the geometric word entity set. Formally, our problem can be articulated as follows: given an undirected weighted graph $G=(V,E)$ and the node feature matrix $X\in \mathbb{R}^{N*K}$, wherein each node is represented by an $N$-dimensional feature vector and $K$ signifies the number of entities in $G$, our objective is to learn embeddings for all nodes within the graph. The final node embedding in graph $G$ is denoted as $X^\in \mathbb{R}^{E*K}$, where each node boasts an E-dimensional embedding. Our ultimate optimization objective can be expressed as follows:
\begin{equation}
min \sum_{v_i,v_j\in V}-log p(1|v_i,v_j)+log p(0|v_i,v_j), s.t~v_i\not= v_j,
\label{objectfunction2}
\end{equation}
Here, $p(1|v_i,v_j)$ signifies the presence of a direct edge between nodes $v_i$ and $v_j$ in graph $G$, while $p(0|v_i,v_j)$ indicates the absence of an edge between them. The objective aims to minimize the direct distance between similar nodes. In scenarios where specific category information about the 3D shape or 2D image is available, additional category edges are introduced between relevant entities. These edges serve to reduce the distance between embeddings of two entities. Furthermore, if a 3D shape and a 2D image share the same geometric word entity, they are connected via a pathway consisting of multiple edges. This information, integrated into the graph embedding process, mitigates disparities between entities, ultimately influencing the final entity embeddings.

In summary, our approach involves generating embeddings for nodes within the 3D shape knowledge graph. This is achieved by utilizing a novel graph convolutional network that leverages the graph's structural characteristics, particularly the geometric word entity, to enhance the consistency of feature vectors for similar 3D shapes and establish meaningful connections between images and 3D shapes. The objective function minimizes the direct distance between related nodes and incorporates category edges and geometric word entities to facilitate effective cross-domain retrieval.

\subsubsection{Graph Neural Network}
As highlighted in the literature, graph neural networks (GNNs) have emerged as a potent and contemporary approach for acquiring representations within graph structures \cite{hamilton2017inductive}. Consequently, we place a pronounced emphasis on the application of Graph Convolutional Networks (GCNs) for embedding the 3D shape knowledge graph. Our primary objective revolves around addressing the challenges associated with cross-domain 3D shape retrieval. In this context, the utilization of GCNs is specifically directed towards enhancing the representation learning process for both shape entities and image entities. It is pertinent to note that these distinct entities inherently possess diverse representations and interact with different sets of neighbors.

Within the context of our knowledge graph, an essential principle is that images and 3D shapes representing the same object should share corresponding geometric entities. This underlines the importance of employing geometric entities to facilitate the generation of embeddings for both image and shape entities. By doing so, we aim to ensure that entities corresponding to identical objects exhibit embeddings that are coherent and comparable. In essence, our approach underscores the pivotal role of Graph Convolutional Networks (GCNs) in addressing the intricacies of cross-domain 3D shape retrieval. This entails leveraging shared geometric entities to underpin the embedding process for image and shape entities, thus establishing a robust and unified foundation for effective representation learning within the 3D shape knowledge graph.

The classic GCN structure\cite{hamilton2017inductive} is used to learn the embeddings of the nodes, which is defined as follows:
\begin{equation}
\begin{split}
{y^{(l+1)}}& = \emph{Relu}({\widetilde{D}}^{-\frac{1}{2}}\widetilde{A}{\widetilde{D}}^{-\frac{1}{2}}y^{(l)}W^{(l)}),
\label{GCN_general}
\end{split}
\end{equation}
where $y^{(l+1)} \in \mathbb{R}^{N\times{F}}$ is the final entity feature matrix in graph $G$. The subscript $l$ is the index of the domain-shared layer. The term $N$ represents the entity number on the knowledge graph. Let $A$ be the original adjacency matrix. The graph adjacency matrix is calculated as $\widetilde{A} = A+I_N \in{\mathbb{R^{N\times{N}}}}$. The learned weight of the $l$-th GCN layer is $W^{(l)}$. Finally, the embeddings $y^{(l+1)}$ is generated by combining all the entity embeddings. The adjacency matrix $\widetilde{D} \in{\mathbb{R^{N\times{N}}}}$ is calculated as follows.
\begin{equation}
\label{adaject3}
\widetilde{D}(i,j) = \sum_{j} \widetilde{A}(i,j).
\end{equation}

To learn the embeddings of an entity, we train the neural network by modeling the graph structure. We use the embeddings to generate the linkage between two entities, i.e., the probability that there exists an edge between $v_j$ and $v_i$ in Eq.\ref{objectfunction2}. Therefore, we formulate embedding learning as a binary classification problem by using the embeddings of two entities.

The probability that there exists an edge between node $v_i$ and node $v_j$ in graph $G$ is defined in Eq.\ref{eq1} and the probability that there exists no edge between node $v_i$ and node $v_j$ in graph $G$ is defined in Eq.\ref{eq2}.
\begin{equation}
p(1|v_i,v_j)=\sigma(y_i^T\dot y_j),
\label{eq1}
\end{equation}
\begin{equation}
p(0|v_i,v_j)=\sigma(-y_i^T\dot y_j),
\label{eq2}
\end{equation}
where $y_i$ is the embedding vector of $v_i$, and $\dot y_j$ is the embedding vector of $v_j$. $\sigma(\dot)$ is the sigmoid function.

Accordingly, the optimization objective function Eq.\ref{objectfunction2} is expressed as:
\begin{equation}
\begin{aligned}
&min \sum_{v_i,v_j\in V}-log p(1|v_i,v_j)+log p(0|v_i,v_j)\\
=&min-\sum_{v_j\in S_p}\log p(1|v_i,v_j)+\sum_{v_k\in S_n}\log p(0|v_i,v_k)\\
=&min-\sum_{v_j\in S_p}\log \sigma(y_i^T.y_j)+\sum_{v_k\in S_n}\log \sigma(y_i^T.y_k),
\end{aligned}
\end{equation}
where $S_p$ is the set of entities in graph $G$ that has a clear pathway to node $v_i$, and $S_n$ is the set of nodes that does not have a pathway to node $v_i$.
Thus, the final objective function is defined as:
\begin{equation}
L=min-\sum_{v_j\in S_p}\log \sigma(y_i^T.y_j)+\sum_{v_k\in S_n}\log \sigma(y_i^T.y_k).
\end{equation}

\subsubsection{Optimization}
In our model, we need to find the optimized parameter $W$. The classic back-propagation algorithm is utilized to optimize this parameter. Thus, the optimization objective is defined as follows:
\begin{equation}
\begin{aligned}
W^* &=\arg \min_{W}L(W)\\
&=\arg \min_{W}-\sum_{v_j\in S_p}\log \sigma(y_i^T\dot y_j)+\sum_{v_k\in S_n}\log \sigma(y_i^T\dot y_k).
\end{aligned}
\end{equation}

The goal of this objective function is to find the solution that is optimal for each entity representation. $W$ is trained by the optimizer according to the gradient:
$W'=W-b\frac{\partial L(W)}{\partial W}$.
Here, we summarize the learning procedure of our approach in Algorithm 1. We first input the graph $G$, the input feature $X$ and the initialization parameter $W$. Then, we sample the training samples $S$. $v_i$ and $v_j$ are sampled from $S$. The existing pathways are utilized for the training of the network, compute the gradients and update the parameters of the specific graph convolutional layers. Finally, we return a set of embeddings for each entity.
\begin{algorithm}[t]
\caption{Graph Convolutional Network}
\hspace*{0.02in} {\bf Input:}
Graph $G=(V,E)$, the node feature matrix $X\in \mathbb{R}^{N*M}$, and the initialized parameter $W$.\\
\hspace*{0.02in} {\bf Output:}
The learned parameter $W$ and the embeddings $Y$.
\begin{algorithmic}[1]
\State Initialize parameters $W$,
\For{$(v_i,v_j)\in$ S}
\State Sample a set of samples $S$.
\State Compute Gradients: $\frac{\partial L(W)}{\partial W}$.
\State Update $W'=W-b\frac{\partial L(W)}{\partial W}$.
\EndFor
\State \Return The final embeddings: $Y$.
\end{algorithmic}
\end{algorithm}
\begin{figure*}[ht]
	\centering
	\includegraphics[width=1\linewidth]{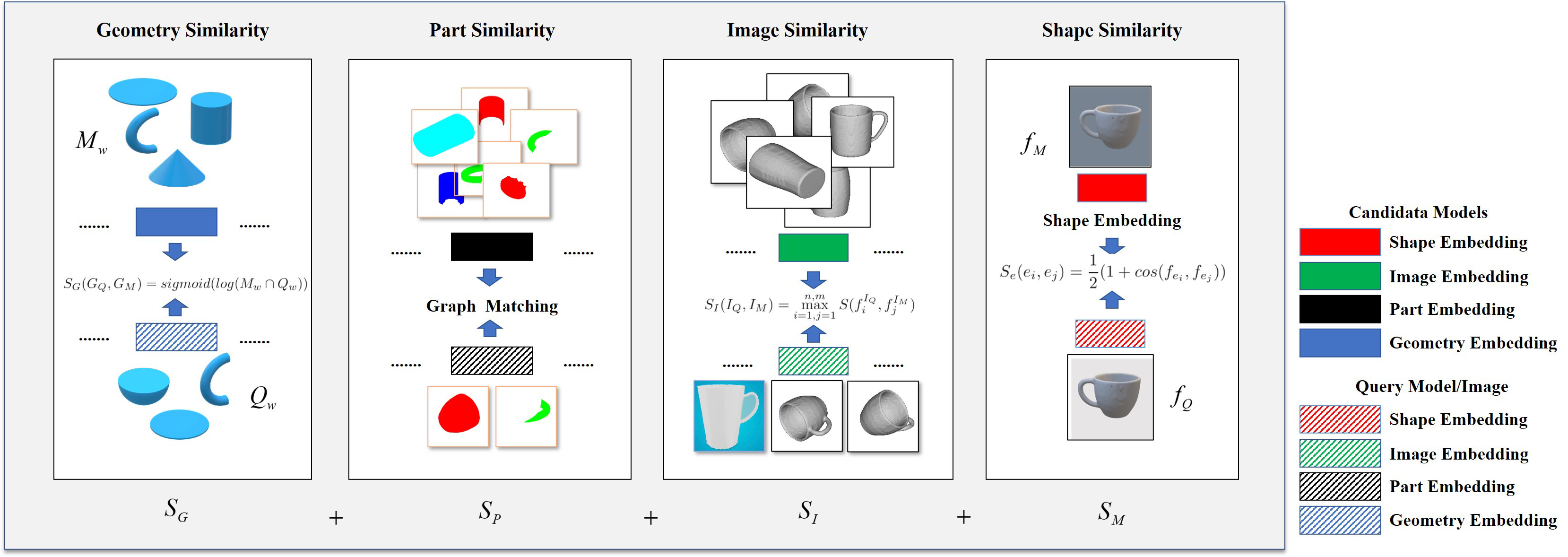}
    \caption{Illustration of the similarity measure. $S_M$ denotes the 3D-shape-level similarity measure, $S_I$ denotes the view-level similarity measure, $S_P$ denotes the part-level similarity measure and $S_G$ denotes the geometry-word-level similarity measure.}
	\label{simialr_figure}
\end{figure*}

\subsection{Similarity Measure}
\label{simeasure}
Based on the above process, a 3D shape knowledge graph and corresponding entity embeddings are generated. The next issue is how to manage the similarity measure between the candidate shapes and the query shape/image according to these embeddings. In this section, we will detail our solution.

Given a query shape $Q$, the query image $I$ and a large 3D shape dataset $\mathbb{M}$, the purpose of 3D shape retrieval is to construct a similarity measure function to calculate the similarity between query shape $Q$, query picture $I$ and candidate shape$M\in \mathbb{M}$. In this work, we first build the 3D shape knowledge graph on the dataset $\mathbb{M}$, query shape $Q$ and query image $I$. Then, the GCN model is utilized to generate the related embeddings for each entity. Finally, we measure the similarity between $Q/I$ and $M$ according to these embeddings.

For each query shape $Q$, the directly related entities include image entities, part entities and geometric entities, as shown in Fig.\ref{demo}. These entities can be represented by a set of embeddings, as shown in Fig.\ref{simialr_figure}. Here, we utilize $I_Q$ to represent the image entity set, $P_Q$ to represent the part entity set and $G_Q$ to represent the geometric word entity set. Each entity will be represented by an embedding. Here, $f$ is utilized to represent the embedding of each entity. Thus, for the query shape $Q$, the image entity set can be represented as $I_Q=\{f_1^{I_Q},f_2^{I_Q},...,f_n^{I_Q}\}$, the part entity set can be represented as $P_Q=\{f_1^{P_Q},f_2^{P_Q},...,f_m^{P_Q}\}$, and the geometric word entity set is represented as $G_Q=\{f_1^{G_Q},f_2^{G_Q},...,f_l^{G_Q}\}$.
Each candidate shape $M$ can also be represented by the related image entity, part entity and geometric entity. Here, we utilize $I_M$, $P_M$ and $G_M$ to represent the three kinds of entity sets. Meanwhile, $I_M=\{f_1^{I_M},f_2^{I_M},...,f_n^{I_M}\}$, $P_M=\{f_1^{P_M},f_2^{P_M},...,f_n^{P_M}\}$ and $G_M=\{f_1^{G_M},f_2^{G_M},...,f_n^{G_M}\}$ are utilized to represent the embeddings of the candidate shapes. The objective is to calculate how similar $Q$ and $M$ are. We first define the similarity between two different entities by Eq.\ref{cossimilarity}.
\begin{equation}
S_e(e_i,e_j)=\frac{1}{2}(1+cos(f_{e_i},f_{e_j})),
\label{cossimilarity}
\end{equation}
where $f_{e_i}$ and $f_{e_j}$ are the embedding features of entities $e_i$ and $e_j$. Here, the classic cosine distance is utilized to compare different entities. The cosine distance has the range $[-1,+1]$. Eq.\ref{cossimilarity} is utilized for normalization.

Then, according to different entities, the similarity between query shape $Q$ and candidate shape $M$ is defined as:
\begin{equation}
\begin{aligned}
    S(Q,M)=&\alpha S_M(f_Q,f_M)+\beta S_I(I_Q,I_Q)\\
    &+\gamma S_P(P_Q,P_M)+\lambda S_G(G_Q,G_M),\\
    s.t. ~~&\alpha+\beta+\gamma+\lambda=1,\\
\end{aligned}
\end{equation}
where $f_Q$ and $f_M$ are the embeddings of the query shape entity and the candidate shape entity, respectively. $S_M(f_Q,f_M)=S_e(f_Q,f_M)$. Meanwhile,
\begin{equation}
S_I(I_Q,I_M)=\max_{i=1,j=1}^{n,m}{S(f_i^{I_Q},f_j^{I_M})},
\end{equation}
where $n$ and $m$ are the number of image entities in $I_Q$ and $I_M$, respectively.
\begin{equation}
S_G(G_Q,G_M)=sigmoid(log(M_w\cap Q_w)).
\label{gemotrys}
\end{equation}
For $S_P(G_Q,G_M)$, we adopt the bipartite graph matching method \cite{gao20113d} for the similitude measurement between different part entity sets. However, the results of graph matching are a set of matching pairs. The matching score is the summary of matching pairs.
Here, we normalize the results by combining averages.
$\alpha$, $\beta$, $\gamma$ and $\lambda$ are the weights of these four similarity measures, which are used to control the contributions of different entities in the measurement.

In other words, for each query image $I$, the directly related entities include part entities and geometric word entities. These entities can be represented by a set of embeddings. Here, $P_I$ and $G_I$ are utilized to represent the part entity set, and the geometric word entity set, respectively. We also utilize $P_I=\{f_1^{I_Q},f_2^{I_Q},...,f_n^{I_Q}\}$ and $G_I=\{f_1^{I_Q},f_2^{I_Q},...,f_n^{I_Q}\}$ to represent the embeddings of these two kinds of entities. The final similitude between the input image and its relevant shape can be calculated as:
\begin{equation}
\begin{aligned}
    S(I,M)=&\beta^* S_I(f_I,I_Q)\\
    &+\gamma^* S_P(P_I,P_M)+\lambda^* S_G(G_I,G_M),\\
    s.t. ~~&\beta^*+\gamma^*+\lambda^*=1,
\end{aligned}
\label{queryimage}
\end{equation}
where $f_I$ is the input query (image) embedding. $I_Q$ is the candidate 3D shape's view embedding set.
\begin{equation}
S_I(I,I_M)=\max_{j=1}^{m}{S(f_I,f_j)},
\end{equation}
where $m$ is the number of views extracted from the candidate 3D shape $M$. $S_G(G_Q,G_M)$ is computed by Eq.\ref{gemotrys}. For $S_P(P_Q,P_M)$, we applied a computational approach similar to shape retrieval. $\beta^*$, $\gamma^*$, and $\lambda^*$ are the weights of these three similarity measures in Eq. \ref{queryimage}, which are also used to balance the contributions of different entities in the final similarity measure.

So far, we are able to handle the 3D shape retrieval and cross-domain 3D shape retrieval without massive training and parameter debugging issues. We only focus on how to enlarge the size of the geometric words in the proposed knowledge graph.

\section{Experiment and Discussion}
We have conducted a comprehensive set of experiments to showcase the efficacy of our proposed approach. Section \ref{retrieval} presents the experimental outcomes pertaining to the conventional 3D shape retrieval task, utilizing the widely-recognized ModelNet40 dataset. In Section \ref{crossdomain}, our focus shifts towards evaluating the performance in the context of cross-domain retrieval. In this scenario, we utilize 3D shapes extracted from the ModelNet40 test set as queries, seeking similar 3D shapes from the Shapenet-Core-55 dataset \cite{Wu20153D}.
Furthermore, we delve into the realm of cross-modal retrieval in Section \ref{SHREC2019}, where we present our experimental investigations. In contrast to the aforementioned experiments, this scenario involves utilizing an input image as the retrieval query, with the corresponding 3D shapes serving as the retrieval targets.
The subsequent subsections will provide a comprehensive breakdown of these experiments, offering intricate insights into each facet of our evaluation process.

\subsection{Dataset}
To evaluate the efficiency of our proposed method, we made considerable use of ModelNet40\cite{Wu20153D} dataset, which contains 12,311 CAD models divided into 40 categories. ModelNet40's training and testing data are made up of 9,843 and 2,468 3D models, respectively. This dataset is utilized to find the best parameters of our model.
To demonstrate our approach on the cross-domain information retrieval task, we use the ShapeNet-Core-55 dataset. This dataset has been used for the Shape Retrieval Contest (SHREC) 2018 competition track to evaluate the performance of 3D shape retrieval methods.
Our method for cross-modal 3D shape retrieval is also shown using the MI3DOR dataset from SHREC 2018. This dataset, which includes 21,000 2D monocular photos of 21 categories and 7,690 3D shapes, is a public benchmark for 3D shape retrieval using monocular images that was provided by \cite{li20193Dor}. The benchmark is split into two sets: a training set that comprises 3,842 3D shapes and 10,500 2D images, and a testing set that uses the remaining data.
We have conducted extensive experiments to evaluate the performance of our method. However, due to the length constraints, we only select some key experiments to describe in the manuscript. More experiments are shown in supplemental files.

\subsection{Evaluation Metrics}
In the context of 3D shape retrieval, the credibility of our experimental outcomes hinges upon the thoroughness of our evaluation methodology. To this end, we employ a range of well-established metrics \cite{liu2017view}, including but not limited to nearest neighbor (NN), first tier (FT), second tier (ST), F-measure (F), and discounted cumulative benefit (DCG). By applying these metrics, we rigorously assess the performance of our approach against state-of-the-art methods, facilitating a comprehensive and insightful comparative analysis.

\subsection{3D shape representation on ModelNet40}
\label{retrieval}
In order to validate the efficacy of our proposed method, we conducted comprehensive 3D shape retrieval experiments utilizing the Princeton ModelNet40 dataset \cite{Wu20153D}. Our comparisons encompass a diverse array of methodologies, spanning different representations of 3D data. These include volumetric-based techniques \cite{Wu20153D}, handcrafted features tailored for multi-view data \cite{SPH} \cite{LFD}, deep learning approaches designed for multi-view data \cite{su2015multi} \cite{MVCNN-MultiRes}, deep learning methods tailored for panorama views \cite{panoramann}, as well as point cloud-based methodologies \cite{qi2017pointnet} \cite{qi2017pointnet++} \cite{wang2018dynamic}. Through these extensive comparisons, we aim to establish a robust assessment of our approach's performance.

\begin{table}[ht]
\centering
\caption{{Experimental results of 3D shape classification and retrieval on ModelNet40.}}\label{model40}
\begin{tabular}{lcc}
    \toprule
    Method                             & Classification (ACC) & Retrieval (mAP) \\
    \midrule
    SPH\cite{SPH}                      & 68.2\%               & 33.3\%          \\
    LFD\cite{chen2003visual}           & 75.5\%               & 40.9\%          \\
    3D ShapeNets\cite{Wu20153D}        & 77.3\%               & 49.2\%          \\
    VoxNet\cite{Allen2008VoxNet}       & 83.0\%               & -               \\
    VRN\cite{VRN}                      & 91.3\%               & -               \\
    MVCNN (AlexNet)\cite{su2015multi}, & 89.5\%               & 80.2\%          \\
    MVCNN (GoogLeNet),                 & 92.2\%               & 83.0\%          \\
    LMVCNN-VggNet-11\cite{yulatent},   & 93.5\%               & -               \\
    VS-MVCNN\cite{ma2017boosting}      & 90.9\%               &                 \\
    PointNet++\cite{qi2017pointnet++}  & 90.7\%               & -               \\
    DGCNN\cite{wang2018dynamic}        & 92.2\%               & -               \\
    PVNet\cite{you2018pvnet}           & 93.2\%               & 89.5\%          \\
    N-gram Network\cite{he2019view}    & 90.2\%               & 89.3\%          \\
    PCT\cite{guo2021pct}               & 93.2\%               & -               \\
    Ours (AlexNet), 12$\times$         & 93.7\%               & 91.8\%          \\
    MLVACN\cite{gao2021multi}          & -                    & \textbf{93.5\%} \\
    Ours (ResNet), 12$\times$          & \textbf{96.9\%}      & 92.7\%          \\
    \bottomrule
\end{tabular}
\end{table}

In our study, we utilized the test data as the query models, while the training data was employed to construct the 3D shape knowledge network. The comparative evaluation against state-of-the-art methods is summarized in Table \ref{model40}. In the context of the 3D classification task, our approach demonstrates superior performance. Notably, our approach outperforms Point Cloud Transformer (PCT)\cite{guo2021pct} by a margin of $3.7\%$. Additionally, in addressing the retrieval problem, our approach also achieves favorable results. Remarkably, in contrast to traditional methods, our approach circumvents the need for an intricate training process, the benefits of which will be exhibited in the next subsection.

\subsection{3D shape retrieval based on cross-domain datasets} \label{crossdomain}
The 3D shape knowledge graph proves to be a highly effective solution for addressing the cross-domain 3D shape retrieval challenge. A characteristic form of this cross-domain predicament involves learning from data that adheres to dissimilar distributions. To elucidate this concept, we employ 3D shapes from the ModelNet40 testing set as query models, seeking related 3D shapes from the ShapeNet-Core-55 dataset. The ensuing experimental results substantiate the efficacy and resilience of the 3D shape knowledge graph, as exemplified in Table \ref{model401-train}. For the purpose of comparison, we have implemented several standard cross-domain learning methodologies\cite{conf/icml/GaninL15,conf/icml/LongZ0J17,conf/cvpr/ZhangLO17,long2014transfer,long2013transfer}. These approaches involve mapping cross-domain embeddings onto an intermediary domain and subsequently gauging their similarity using the Euclidean distance metric.

Significantly, our approach dispenses with the need for training, setting it apart from these cross-domain learning methods. In this context, we directly evaluated our performance without any training or fine-tuning interventions. As evidenced by the data presented in Table \ref{model401-train}, the disparity in performance between state-of-the-art (SOTA) methods and our approach is negligible. Nonetheless, even though our method is applied to the cross-domain dataset without training, it is still successful in addressing the cross-domain retrieval task and is comparable to or outperforms other state-of-the-art methods.

\begin{table}[ht]
\centering
\caption{{Experimental results on the cross-domain dataset (trained on this task). Queries are from ModelNet40 and candidates are retrieved from ShapeNet.}}\label{model401-train}
\begin{tabular}{lcccccc}
    \toprule
    Method                           & NN            & FT            & ST            & F             & DCG           & ANMRR         \\
    \midrule
    RevGard\cite{conf/icml/GaninL15} & 0.89          & 0.79          & 0.90          & 0.33          & 0.83          & 0.15          \\
    JAN\cite{conf/icml/LongZ0J17}    & 0.90          & 0.81          & 0.91          & 0.33          & 0.84          & 0.14          \\
    TJM\cite{long2014transfer}       & 0.90          & 0.81          & 0.91          & 0.34          & 0.84          & 0.14          \\
    JDA\cite{long2013transfer}       & 0.91          & 0.82          & 0.91          & 0.34          & 0.85          & 0.14          \\
    JGSA\cite{conf/cvpr/ZhangLO17}   & 0.92          & 0.82          & 0.92          & 0.34          & 0.85          & 0.13          \\
    KGMR\cite{nie20203d}             & 0.91          & 0.83          & 0.92          & 0.34          & 0.87          & 0.13          \\
    IPSC\cite{song2023self}          & 0.93          & 0.83          & 0.93          & \textbf{0.36} & 0.87          & 0.12          \\
    SI3DMR\cite{li2023instance}      & \textbf{0.94} & \textbf{0.84} & 0.93          & 0.35          & \textbf{0.88} & \textbf{0.10} \\
    Ours                             & 0.93          & \textbf{0.84} & \textbf{0.94} & \textbf{0.36} & 0.87          & 0.12          \\
    \bottomrule
\end{tabular}
\end{table}

\subsection{3D shape retrieval on cross-modal conditions}
\label{SHREC2019}
The cross-domain situations are currently narrowed as cross-modal conditions in 3D shape retrieval tasks. To verify the performance of the proposed method on the cross-modality information retrieval, we also conduct experiments on a dataset more suitable for this task, SHREC 2019 Monocular Image-Based 3D Object Retrieval (MI3DOR)~\cite{li20193Dor}.
The comparison methods on the MI3DOR dataset involve both supervised and unsupervised methods. They are distinguished by whether labels are available for the target domain data. We compare our approach with both kinds of these methods. The related experimental results are listed in Table.\ref{supervised} for supervised methods and Table.\ref{unsupervised_result} for unsupervised methods. According to the results, our method achieves promising results. The results demonstrate the performance of various methods on the MI3DOR dataset, focusing on cross-modal retrieval, which is similar to image-based 3D shape retrieval. In the supervised methods, our method consistently outperforms other techniques across most metrics. It achieves the highest retrieval rates, indicating its effectiveness in retrieving relevant 3D shapes.

The unsupervised methods also highlight the competitiveness of the proposed method. While it may not achieve the highest values in all metrics, it still maintains strong performance across the board. Notably, in the unsupervised setting, our method performs comparably or better than most other methods, and it achieves the best F\_measure, indicating a better ranking of retrieved shapes.

Overall, the experimental results suggest that the proposed method, which leverages a 3D shape knowledge graph with integrated geometry information, yields favorable cross-modal retrieval performance in both supervised and unsupervised scenarios on the MI3DOR dataset. This integration of geometry-based knowledge into the knowledge graph seems to contribute positively to the retrieval of relevant 3D shapes, making the proposed method a promising approach for 3D shape retrieval tasks.

\begin{table}[ht]
\centering
\caption{{Experimental results of supervised methods on MI3DOR dataset. All methods for comparison are from SHREC 2019 Monocular Image-Based 3D Object Retrieval (MI3DOR)~\cite{li20193Dor}}}\label{supervised}
\begin{tabular}{lccccccc}
    \toprule
    {Method}     & {NN}          & {FT}          & {ST}          & {F}           & {DCG}         & {ANMRR}       \\ \midrule
    {RNF-MVCVR}  & 0.97          & 0.92          & 0.93          & 0.2           & 0.93          & 0.06          \\
    {SORMI}      & 0.94          & 0.92          & 0.96          & 0.18          & 0.92          & 0.07          \\
    {RNFETL}     & 0.97          & 0.91          & 0.97          & 0.18          & 0.92          & 0.07          \\
    {CLA}        & 0.95          & 0.88          & 0.89          & 0.2           & 0.9           & 0.1           \\
    {MLIS}       & 0.94          & 0.91          & 0.96          & 0.18          & 0.91          & 0.08          \\
    {ADDA-MVCNN} & 0.87          & 0.86          & 0.87          & 0.17          & 0.87          & 0.13          \\
    {SRN}        & 0.89          & 0.86          & 0.87          & 0.18          & 0.88          & 0.12          \\
    {ALIGN}      & 0.64          & 0.69          & 0.8           & 0.13          & 0.69          & 0.3           \\
    {Ours}       & \textbf{0.98} & \textbf{0.93} & \textbf{0.98} & \textbf{0.21} & \textbf{0.94} & \textbf{0.06} \\ \bottomrule
\end{tabular}
\end{table}

\begin{table}[ht]
\centering
\caption{{Experimental results of unsupervised methods on MI3DOR dataset.}}\label{unsupervised_result}
\begin{tabular}{lcccccc}
\toprule
    {Method}                            & {NN}          & {FT}          & {ST}          & {F}           & {DCG}         & {ANMRR}       \\ \midrule
    MEDA\cite{wang2018visual}           & 0.43          & 0.34          & 0.50          & 0.05          & 0.36          & 0.65          \\
    {JMMD-AlexNet}\cite{li20193Dor}     & 0.44          & 0.34          & 0.49          & 0.08          & 0.364         & 0.64          \\
    JAN\cite{conf/icml/LongZ0J17}       & 0.45          & 0.34          & 0.50          & 0.09          & 0.36          & 0.65          \\
    {MVML}\cite{li20193Dor}             & 0.61          & {0.44}        & {0.59}        & {0.11}        & 0.47          & {0.54}        \\
    JGSA\cite{conf/cvpr/ZhangLO17}      & 0.61          & 0.44          & 0.60          & 0.12          & 0.47          & 0.54          \\
    IPSC\cite{song2023self}             & 0.73          & \textbf{0.65} & \textbf{0.81} & 0.15          & \textbf{0.67} & \textbf{0.34} \\
    SI3DMR\cite{li2023instance}         & \textbf{0.78} & 0.58          & 0.73          & 0.15          & 0.62          & 0.40          \\
    {Ours}                              & 0.64          & 0.47          & 0.62          & \textbf{0.17} & 0.53          & 0.49          \\ \bottomrule
\end{tabular}
\end{table}

\begin{figure*}[htbp]
	\centering
	\includegraphics[width=1\linewidth]{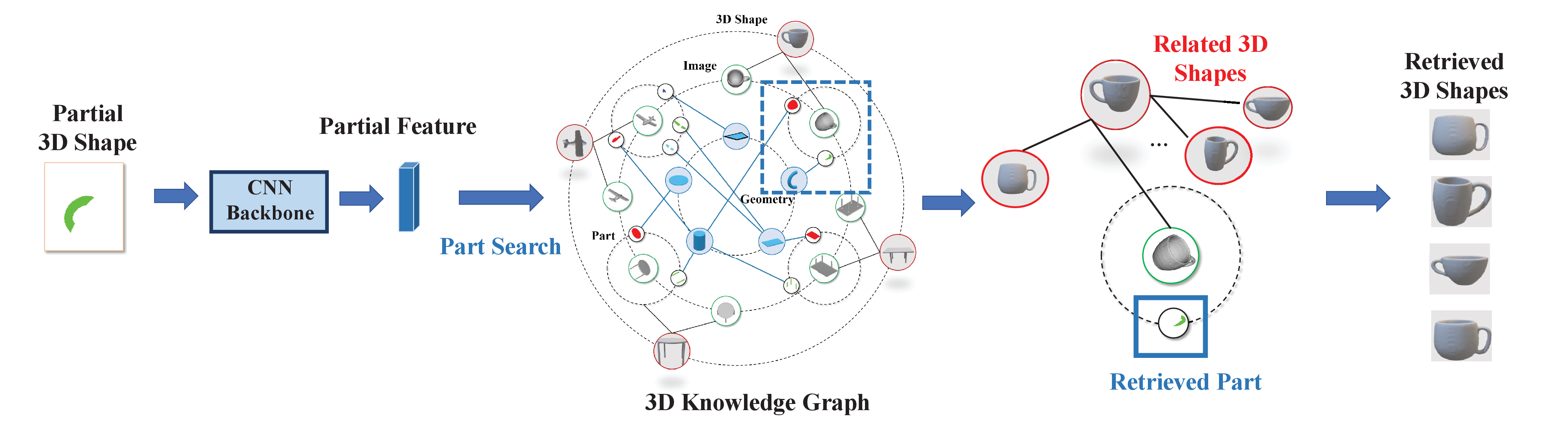}
    \caption{Partial 3D model retrieval framework.}
	\label{ap1}
\end{figure*}

\begin{figure}[hbtp]
	\centering
	\includegraphics[width=1\linewidth]{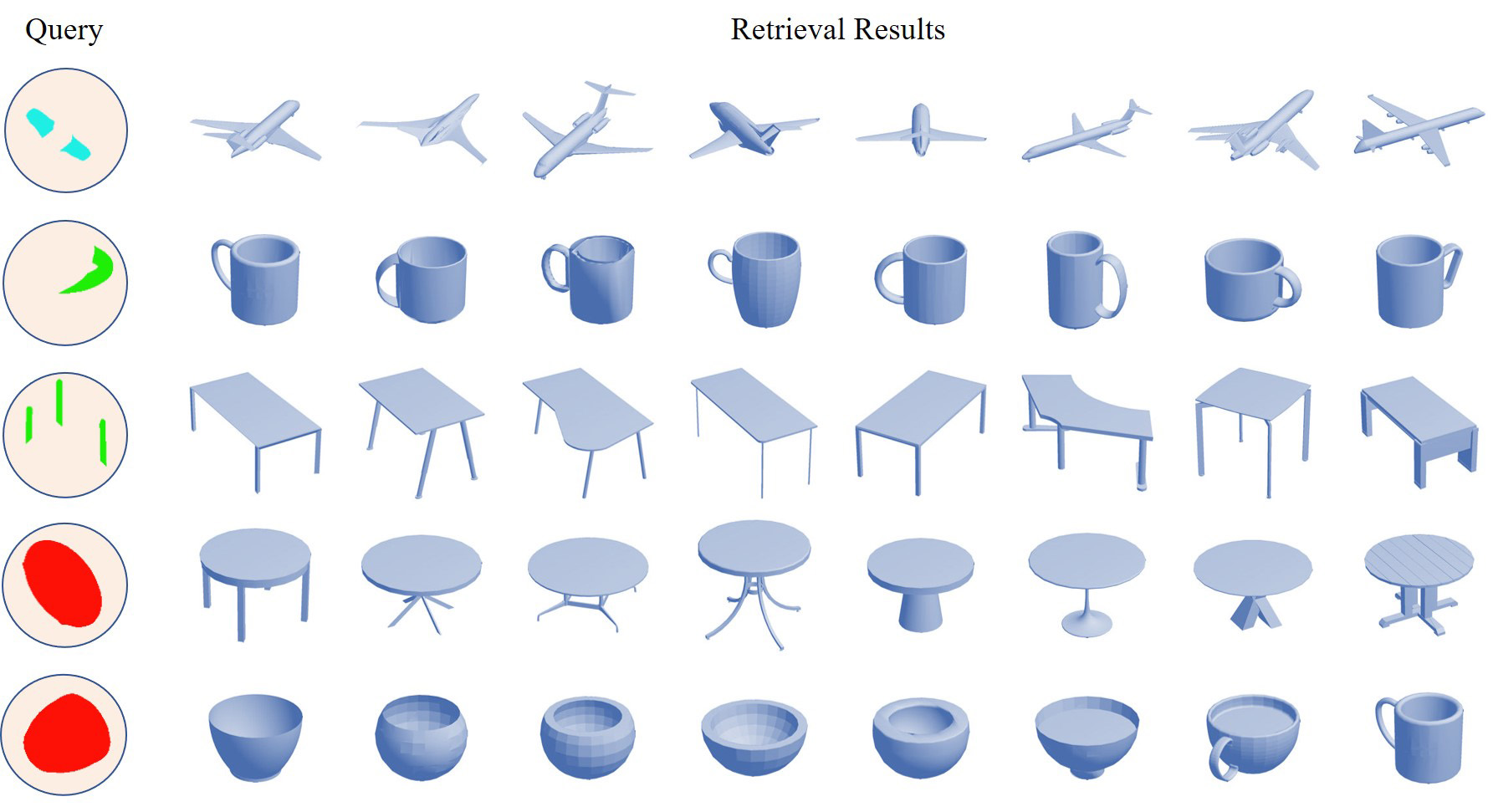}
	\caption{Partial 3D model retrieval samples. (Zoom in for more details.)}
	\label{partial_retrieval}
\end{figure}

\begin{figure*}[htp]
	\centering
	\includegraphics[width=1\linewidth]{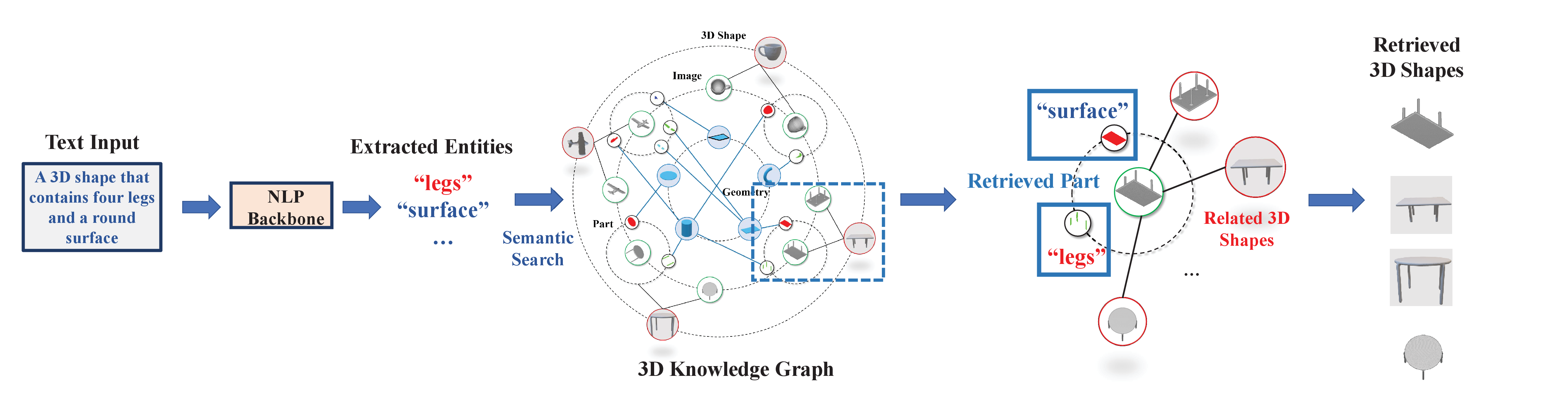}
    \caption{Text-3D model retrieval framework.}
	\label{ap2}
\end{figure*}

\begin{figure}[htp]
	\centering
	\includegraphics[width=1\linewidth]{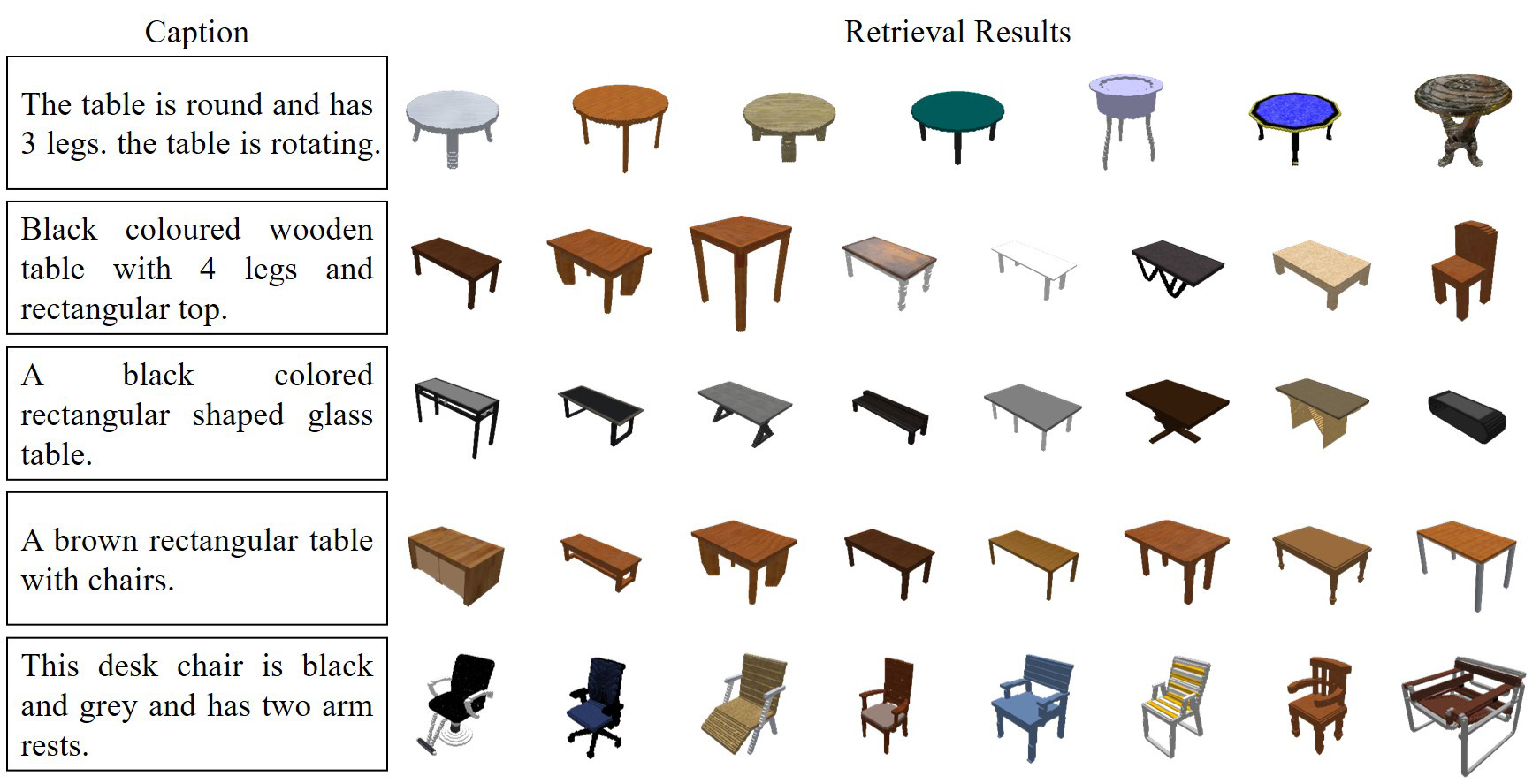}
	\caption{Text-3D model retrieval samples.(Zoom in for more details.)}
	\label{text-to-3D}
\end{figure}

\subsection{Discussion on Similarity Measurement}
In this section, we present a comprehensive similarity calculation method that encompasses various entities, as illustrated in Fig.\ref{simialr_figure}. To evaluate the efficacy of our proposed approach, we conducted a thorough comparative analysis involving different similarity measures, using the ModelNet40 dataset as our experimental platform. The results of these experiments are summarized in Table.\ref{similarity}.

\begin{table*}[ht]
    \centering
    \caption{{Comparison of different similar measures on ModelNet40.}}\label{similarity}
    \begin{tabular}{ccccccccc}
        \toprule
        Supervision & Layer & NN    & FT    & ST    & F\_measure    & DCG   & ANMRR \\ \midrule
        \multirow{5}{*}{Unsuperviesd}            & $S_G$ & 0.84  & 0.67  & 0.82  & 0.27          & 0.70  & 0.27	 \\
                    & $S_P$ & 0.83  & 0.66  & 0.81  & 0.25          & 0.69  & 0.28 \\
                    & $S_I$ &0.82 &0.65 &0.80 &0.24 &0.68 &0.29 \\
                    & $S_M$&0.83 &0.66 &0.81 &0.26 &0.69 & 0.28 \\
                    & $S_G+S_P+S_I+S_M$ &\textbf{0.85} &\textbf{0.69} &\textbf{0.84} &\textbf{0.30} &\textbf{0.73} &\textbf{0.26} \\
        \midrule
        \multirow{5}{*}{Superviesd}    & $S_G$ &0.87 &0.78 &0.88 &0.31 &0.81 & 0.17	\\
                            & $S_P$ &0.89 &0.80 &0.90 &0.32 &0.83 &0.15 \\
                            & $S_I$&0.88&0.80 &0.90 &0.31 &0.82 & 0.16 \\
                            & $S_M$&0.90 &0.81 &0.91 &0.33 &0.84 & 0.15 \\
                            & $S_G+S_P+S_I+S_M$&\textbf{0.92} &\textbf{0.83} &\textbf{0.93} &\textbf{0.35} &\textbf{0.86} &\textbf{0.13} \\
        \bottomrule
    \end{tabular}%
    \end{table*}%

From the experimental outcomes, it is evident that the supervised method outperforms the unsupervised alternatives. This superiority can be attributed to the inclusion of the category edge within the 3D shape knowledge graph. The additional information introduced by the category edge contributes to the enhancement of the embedding quality produced by the GCN method. In other words, our amalgamated similarity measure achieves the most favorable retrieval results. However, among the single similarity measures, $S_M$ stands out by yielding the most promising outcomes. Nevertheless, in the context of unsupervised methods, $S_G$ manages to achieve a slight improvement over other single similarity measures. This discrepancy can be attributed to the reliance of the graph embedding method on category information during the supervised learning stage. The category edge establishes a clear path between relevant entities, thereby fortifying their relationships. Consequently, $S_P$, $S_I$, and $S_M$ contribute more information to the similarity measure. In unsupervised learning scenarios, the absence of the category edge in the 3D shape knowledge graph places greater emphasis on the geometric entity as the sole bridge between disparate shapes and different modalities. This entity assumes a pivotal role in the graph embedding process, causing the information conveyed by $S_P$, $S_I$, and $S_M$ to converge toward $S_G$, as revealed through the graph embeddings operation. These findings are consistently supported by the final experimental results.

Building upon these insights, we introduce weighted coefficients for each similarity measure in the final step of similarity fusion. Specifically, under the supervised setting, the weights assigned to the four similarity measures ($S_G$, $S_P$, $S_I$, and $S_M$) are 0.1, 0.3, 0.2, and 0.4, respectively, when the query pertains to a 3D shape. When the query involves a 2D image, the weights for the three similarity measures ($S_G$, $S_P$, and $S_I$) are set at 0.2, 0.4, and 0.4, respectively. In the unsupervised scenario, the weights for the four similarity measures ($S_G$, $S_P$, $S_I$, and $S_M$) are adjusted to 0.3, 0.2, 0.2, and 0.3, respectively, for 3D shape queries. Similarly, for 2D image queries, the weights for the three similarity measures ($S_G$, $S_P$, and $S_I$) are fine-tuned to 0.3, 0.2, and 0.4, respectively.

\subsection{Discussion on the Geometric Word Entities}
In this section, we introduce a crucial element of our proposed framework - the geometric word entity, which serves as the cornerstone of our 3D shape knowledge graph and functions akin to a dictionary for object representation. As such, the dimensionality of this dictionary plays a pivotal role in our entire framework and wields a direct impact on the ultimate performance. To shed light on this dimensionality's influence, we conduct comparative experiments with varying numbers of geometric word entities and present the results in Table \ref{numberwords}.

\begin{table*}[ht]
  \centering
  \caption{{Performance on different numbers of geometric word entities on ModelNet40.}}\label{numberwords}
      \begin{tabular}{ccccccccc}
        \toprule
       Supervision & Number Words      & NN    & FT    & ST    & F\_measure & DCG   & ANMRR \\ \midrule
       \multirow{6}{*}{Unsupervised} & 64 &0.76 &0.60 &0.75 &0.22 &0.63 & 0.34	\\
       & 128 &0.79 &0.62 &0.76 &0.24 &0.65 &0.31 \\
       & 256 &0.81 &0.64 &0.79 &0.26 &0.68 &0.29 \\
       & 512 &0.83 &0.67 &0.81 &0.28 &0.70 & 0.27 \\
       & \textbf{1024} &\textbf{0.85} &\textbf{0.69} &\textbf{0.84} &\textbf{0.30} &\textbf{0.73} &\textbf{0.26}  \\
       & 2048 &0.84 &0.67 &0.82 &0.28 &0.71 &0.27 \\
       \midrule
       \multirow{6}{*}{Supervised}& 64 &0.84 &0.73 &0.83 &0.26 &0.76 & 0.21	\\
       & 128 &0.86 &0.77 &0.87 &0.30 &0.80 &0.17 \\
       & 256 &0.89 &0.80 &0.89 &0.32 &0.83 &0.15 \\
       & 512 &0.90 &0.81 &0.91 &0.33 &0.84 & 0.14 \\
       & \textbf{1024} &\textbf{0.92} &\textbf{0.83} &\textbf{0.93} &\textbf{0.35} &\textbf{0.86} &\textbf{0.13} \\
       & 2048 &0.91 &0.82 &0.91 &0.33 &0.84 &0.13 \\
       \bottomrule
        \end{tabular}
\end{table*}

To gauge the efficacy of our approach in the context of the retrieval task, we leverage the ModelNet40 dataset. Observing the experimental outcomes, we discern a trend of improved retrieval results as the number of geometric words increases. Interestingly, we note that the optimal outcome is achieved when the number of geometric words reaches 1024, after which a gradual degradation in performance is observed.

This behavior can be attributed to the introduction of additional geometric words resulting in a sparser structural information landscape. As the number of geometric words grows, there is a possibility of similar parts from different virtual views of a 3D shape being mapped to distinct geometric words. Consequently, the coherence between the two models diminishes, causing the 3D shape knowledge graph to devolve into disjoint clusters. This phenomenon hampers the identification of analogous structural nodes and weakens the impact of the graph embedding process. In essence, the structural relationship information within the 3D shape knowledge graph becomes muddled with a smaller number of geometric words. Conversely, a larger number can lead to the convergence of various part structures onto the same geometric word, thereby diluting the structural information and negatively impacting embedding and retrieval performance.

\subsection{Computational complexity}
In this experiment, we conducted a comparison of the computational complexity of our model with that of several classic 3D representation methods, using the ModelNet40 dataset as a benchmark. We adopted DGCNN \cite{wang2018dynamic} as the reference method for this comparison. The comparison approaches have been selected based on their well-established recognition within the field. These approaches serve as essential benchmarks to establish a foundation for evaluating computational efficiency. The results in Table \ref{computation} presents a comparison of various methods in terms of their computational complexity for 3D shape analysis. Two important aspects are considered: model size and inference time.

Model Size: Smaller model sizes are generally desirable as they require less storage and memory. In this context, Pointnet has the smallest model size of 9.4 MB, followed closely by MVCNN with 11 MB. DGCNN and RotationNet have moderate model sizes of 21 MB and 36 MB, respectively. On the other hand, PointNet++ and PCNN have larger model sizes of 12 MB and 94 MB, respectively. The proposed method has a model size of 51 MB.

Inference Time: Faster inference times are preferred as they lead to quicker predictions. Pointnet demonstrates the lowest inference time of 6.8 ms, followed by MVCNN with 12.3 ms. DGCNN and our method achieve inference times of 27.2 ms and 34 ms, respectively. The other methods, including RotationNet, PointNet++, and PCNN have longer inference times ranging from 117 ms to 163.2 ms.

Overall, the results indicate that our method performs competitively in terms of both model size and inference time compared to the other methods evaluated. We utilized the GCN model to generate node embeddings for retrieval purposes. Our GCN structure comprises only 3 layers, offering distinct advantages in terms of algorithm complexity when compared to traditional deep learning networks. It strikes a balance between model size and inference speed, showcasing promising computational efficiency. While Pointnet and MVCNN exhibit lower inference times, our method demonstrates a reasonable trade-off between model size and inference speed, positioning it as a viable option for efficient 3D shape analysis and recognition.
\begin{table}[ht]
\centering
\caption{Experimental results for the computational complexity, we report the model size and inference time.}\label{computation}
\begin{tabular}{lcc}
\toprule
    Method                                    & Model Size (MB) & Inference Time (ms) \\ \midrule
    Pointnet\cite{qi2017pointnet}             & 9.4             & 6.8               \\
    PointNet++\cite{qi2017pointnet++}         & 12              & 163.2             \\
    PCNN\cite{li2018pointcnn}                 & 94              & 117               \\
    DGCNN\cite{wang2019dynamic}               & 21              & 27.2              \\
    MVCNN\cite{su2015multi}                   & 11              & 12.3              \\
    RotationNet\cite{kanezaki2018rotationnet} & 36              & 121               \\
    Ours                                      & 51              & 34                \\ \bottomrule
\end{tabular}
\end{table}

\section{Extended Applications based on Knowledge Graph}
This paper capitalizes on shape parts to generate geometric words, a pivotal component of our approach. These geometric words serve as building blocks for constructing the 3D shape knowledge graph and facilitating cross-domain3D model retrieval. Our proposed knowledge graph is versatile, extending beyond its primary applications. In this section, we conduct novel experiments to corroborate the effectiveness of the 3D shape knowledge graph in diverse contexts.

\subsection{Partial 3D Shape Retrieval}
The challenge here entails querying with incomplete 3D models. A partial 3D model retrieval system is tasked with returning a ranked list of complete models from a database based on their similarity to the query. Addressing this, we introduce an effective framework, as illustrated in Figure \ref{ap1}. The process involves extracting rendered images from the partial 3D model, followed by feature vector extraction using the ResNet-34 model \cite{DBLP:conf/cvpr/HeZRS16}. These features facilitate the identification of analogous part entities within the 3D knowledge graph. Subsequently, we leverage this information to retrieve the relevant 3D shape based on these part entities. Figure \ref{partial_retrieval} illustrates retrieval examples, substantiating the feasibility of our method.

\subsection{Text-3D Shape Retrieval}
The versatility of the 3D shape knowledge graph extends to text-to-shape retrieval scenarios. For instance, consider the query: "A 3D shape that features four legs and a rounded surface." We initiate the process by employing an NLP method \cite{DBLP:journals/tcsv/PengC20} to extract entities from the given text. These entities are subsequently mapped to corresponding geometric shapes, which are further linked to geometric words or part entities within the 3D knowledge graph. The final step involves selecting 3D shapes with distinct relationships to these geometric words and part entities as the retrieval results. The conceptual framework is depicted in Figure \ref{ap2}, with retrieval instances showcased in Figure \ref{text-to-3D}. We validate this concept using samples from the Text2Shape dataset \cite{chen2018text2shape}, where the construction of the 3D shape knowledge graph and the mapping between geometric shapes and descriptions underscore the practical viability of our approach.

\section{Conclusion}
This study introduces the innovative concept of the "geometric word," a foundational element within our 3D shape knowledge graph. Empirical results underscore the graph's efficacy in capturing and quantifying shape similarities. The "geometric word" concept bridges the gap between 3D shapes across different domains and the interface between 3D shapes and 2D images across varied modalities. Augmenting this notion, our graph embedding strategy harnesses graph structural information and effective similarity measures, making our approach a versatile tool for tackling challenges encompassing 3D shape retrieval and cross-domain 3D shape retrieval.

The empirical findings emphasize the significance of constructing knowledge graphs within the constraints of available data. Furthermore, the adaptability of "geometric words," learnable from a multitude of datasets, points toward the potential for a universal 3D shape knowledge graph akin to WordNet, capable of encompassing a broad spectrum of shape information. Future research will concentrate on realizing this universal 3D shape knowledge graph and exploring its applicability across diverse domains.

% \section{Acknowledgments}
% This work was supported in part by the National Natural Science Foundation of China (31770904, 61872267, 61702471, and 61772359). Natural Science Foundation of Tianjin City (19JCQNJC00500).

\bibliographystyle{vancouver}
\bibliography{Manuscript}

\begin{thebibliography}{10}

\bibitem{9127813}
Guo Y, Wang H, Hu Q, Liu H, Liu L, Bennamoun M.
\newblock Deep learning for 3d point clouds: A survey.
\newblock IEEE transactions on pattern analysis and machine intelligence.
  2020;43(12):4338-64.

\bibitem{9055070}
Wang N, Zhang Y, Li Z, Fu Y, Yu H, Liu W, et~al.
\newblock Pixel2Mesh: 3D mesh model generation via image guided deformation.
\newblock IEEE transactions on pattern analysis and machine intelligence.
  2020;43(10):3600-13.

\bibitem{su2015multi}
Su H, Maji S, Kalogerakis E, Learned-Miller E.
\newblock Multi-view convolutional neural networks for 3d shape recognition.
\newblock In: Proceedings of the IEEE international conference on computer
  vision; 2015. p. 945-53.

\bibitem{DBLP:journals/pr/LeiZZGML19}
Lei Y, Zhou Z, Zhang P, Guo Y, Ma Z, Liu L.
\newblock Deep point-to-subspace metric learning for sketch-based 3D shape
  retrieval.
\newblock Pattern Recognit. 2019;96.

\bibitem{DBLP:conf/jcdl/GoldfederA08}
Goldfeder C, Allen PK.
\newblock Autotagging to improve text search for 3d models.
\newblock In: {ACM/IEEE} Joint Conference on Digital Libraries, {JCDL} 2008,
  Pittsburgh, PA, USA, June 16-20, 2008; 2008. p. 355-8.

\bibitem{qi2017pointnet}
Qi CR, Su H, Mo K, Guibas LJ.
\newblock Pointnet: Deep learning on point sets for 3d classification and
  segmentation.
\newblock In: Proceedings of the IEEE Conference on Computer Vision and Pattern
  Recognition; 2017. p. 652-60.

\bibitem{jiang2019mlvcnn}
Jiang J, Bao D, Chen Z, Zhao X, Gao Y.
\newblock MLVCNN: Multi-loop-view convolutional neural network for 3D shape
  retrieval.
\newblock In: Proceedings of the AAAI conference on artificial intelligence.
  vol.~33; 2019. p. 8513-20.

\bibitem{you2018pvnet}
You H, Feng Y, Ji R, Gao Y.
\newblock Pvnet: A joint convolutional network of point cloud and multi-view
  for 3d shape recognition.
\newblock In: 2018 ACM Multimedia Conference on Multimedia Conference. ACM;
  2018. p. 1310-8.

\bibitem{Dai2018Deep}
Dai G, Xie J, Fang Y.
\newblock Deep Correlated Holistic Metric Learning for Sketch-Based 3D Shape
  Retrieval.
\newblock IEEE Transactions on Image Processing A Publication of the IEEE
  Signal Processing Society. 2018;27(7):3374.

\bibitem{long2014transfer}
Long M, Wang J, Ding G, Sun J, Yu PS.
\newblock Transfer joint matching for unsupervised domain adaptation.
\newblock In: Proceedings of the IEEE conference on computer vision and pattern
  recognition; 2014. p. 1410-7.

\bibitem{Zhang2017Joint}
Zhang J, Li W, Ogunbona P.
\newblock Joint geometrical and statistical alignment for visual domain
  adaptation.
\newblock In: Proceedings of the IEEE conference on computer vision and pattern
  recognition; 2017. p. 1859-67.

\bibitem{Wu20153D}
Wu Z, Song S, Khosla A, Yu F, Zhang L, Tang X, et~al.
\newblock 3d shapenets: A deep representation for volumetric shapes.
\newblock In: Proceedings of the IEEE conference on computer vision and pattern
  recognition; 2015. p. 1912-20.

\bibitem{qi2017pointnet++}
Qi CR, Yi L, Su H, Guibas LJ.
\newblock Pointnet++: Deep hierarchical feature learning on point sets in a
  metric space.
\newblock In: Advances in neural information processing systems; 2017. p.
  5099-108.

\bibitem{DBLP:journals/pami/GarroG16}
Garro V, Giachetti A.
\newblock Scale Space Graph Representation and Kernel Matching for Non Rigid
  and Textured 3D Shape Retrieval.
\newblock {IEEE} Trans Pattern Anal Mach Intell. 2016;38(6):1258-71.

\bibitem{DBLP:journals/pami/XieDZWF17}
Xie J, Dai G, Zhu F, Wong EK, Fang Y.
\newblock DeepShape: Deep-Learned Shape Descriptor for 3D Shape Retrieval.
\newblock {IEEE} Trans Pattern Anal Mach Intell. 2017;39(7):1335-45.

\bibitem{wang2019dynamic}
Wang Y, Sun Y, Liu Z, Sarma SE, Bronstein MM, Solomon JM.
\newblock Dynamic graph cnn for learning on point clouds.
\newblock ACM Transactions on Graphics (tog). 2019;38(5):1-12.

\bibitem{LFD}
Chen DY, Tian XP, Shen YT, Ming O.
\newblock On Visual Similarity Based 3D Model Retrieval.
\newblock Computer Graphics Forum. 2010;22(3):223-32.

\bibitem{GIFT}
{Bai} S, {Bai} X, {Zhou} Z, {Zhang} Z, {Tian} Q, {Latecki} LJ.
\newblock GIFT: Towards Scalable 3D Shape Retrieval.
\newblock IEEE Transactions on Multimedia. 2017 June;19(6):1257-71.

\bibitem{panoramann}
Sfikas K, Pratikakis I, Theoharis T.
\newblock Ensemble of panorama-based convolutional neural networks for 3d model
  classification and retrieval.
\newblock Computers \& Graphics. 2018;71:208-18.

\bibitem{wang2018LocalSpecGCN}
Wang C, Samari B, Siddiqi K.
\newblock Local Spectral Graph Convolution for Point Set Feature Learning.
\newblock In: Computer Vision - {ECCV} 2018 - 15th European Conference, Munich,
  Germany, September 8-14, 2018, Proceedings, Part {IV}. vol. 11208 of Lecture
  Notes in Computer Science. Springer; 2018. p. 56-71.

\bibitem{Te2018RGCNN}
Te G, Hu W, Zheng A, Guo Z.
\newblock Rgcnn: Regularized graph cnn for point cloud segmentation.
\newblock In: Proceedings of the 26th ACM international conference on
  Multimedia; 2018. p. 746-54.

\bibitem{Shi2020Point-GNN}
Shi W, Rajkumar R.
\newblock Point-GNN: Graph Neural Network for 3D Object Detection in a Point
  Cloud.
\newblock In: 2020 {IEEE/CVF} Conference on Computer Vision and Pattern
  Recognition, {CVPR} 2020, Seattle, WA, USA, June 13-19, 2020. {IEEE}; 2020.
  p. 1708-16.

\bibitem{Zhang2021SGG}
Zhang C, Yu J, Song Y, Cai W.
\newblock Exploiting edge-oriented reasoning for 3d point-based scene graph
  analysis.
\newblock In: Proceedings of the IEEE/CVF conference on computer vision and
  pattern recognition; 2021. p. 9705-15.

\bibitem{zhang2022tn}
Zhang L, Chang X, Liu J, Luo M, Li Z, Yao L, et~al.
\newblock Tn-zstad: Transferable network for zero-shot temporal activity
  detection.
\newblock IEEE Transactions on Pattern Analysis and Machine Intelligence.
  2022;45(3):3848-61.

\bibitem{li2023object}
Li Z, Xu P, Chang X, Yang L, Zhang Y, Yao L, et~al.
\newblock When object detection meets knowledge distillation: A survey.
\newblock IEEE Transactions on Pattern Analysis and Machine Intelligence. 2023.

\bibitem{li2022video}
Li M, Huang PY, Chang X, Hu J, Yang Y, Hauptmann A.
\newblock Video pivoting unsupervised multi-modal machine translation.
\newblock IEEE Transactions on Pattern Analysis and Machine Intelligence.
  2022;45(3):3918-32.

\bibitem{he2022category}
He S, Wang W, Wang Z, Xu X, Yang Y, Wang X, et~al.
\newblock Category alignment adversarial learning for cross-modal retrieval.
\newblock IEEE Transactions on Knowledge and Data Engineering.
  2022;35(5):4527-38.

\bibitem{li2018deep}
Li Z, Tang J, Mei T.
\newblock Deep collaborative embedding for social image understanding.
\newblock IEEE transactions on pattern analysis and machine intelligence.
  2018;41(9):2070-83.

\bibitem{li2015weakly}
Li Z, Tang J.
\newblock Weakly supervised deep metric learning for community-contributed
  image retrieval.
\newblock IEEE Transactions on Multimedia. 2015;17(11):1989-99.

\bibitem{mu2018image-based}
Mu P, Zhang S, Zhang Y, Ye X, Pan X.
\newblock Image-based 3D model retrieval using manifold learning.
\newblock Journal of Zhejiang University Science C. 2018;19(11):1397-408.

\bibitem{li2015joint}
Li Y, Su H, Qi CR, Fish N, Cohenor D, Guibas LJ.
\newblock Joint embeddings of shapes and images via CNN image purification.
\newblock international conference on computer graphics and interactive
  techniques. 2015;34(6):234.

\bibitem{han2019seqviews2seqlabels:}
Han Z, Shang M, Liu Z, Vong CM, Liu Y, Zwicker M, et~al.
\newblock SeqViews2SeqLabels: Learning 3D Global Features via Aggregating
  Sequential Views by RNN With Attention.
\newblock IEEE Transactions on Image Processing. 2019;28(2):658-72.

\bibitem{nie20203d}
Nie W, Wang Y, Song D, Li W.
\newblock 3D Model Retrieval Based on a 3D Shape Knowledge Graph.
\newblock IEEE Access. 2020;8:142632-41.

\bibitem{liu2018physical}
Liu Z, Freeman WT, Tenenbaum JB, Wu J.
\newblock Physical primitive decomposition.
\newblock In: Proceedings of the European Conference on Computer Vision (ECCV);
  2018. p. 3-19.

\bibitem{katageri2021pointdccnet}
Katageri S, Kulmi S, Tabib RA, Mudenagudi U.
\newblock PointDCCNet: 3D Object Categorization Network Using Point Cloud
  Decomposition.
\newblock In: Proceedings of the IEEE/CVF Conference on Computer Vision and
  Pattern Recognition; 2021. p. 2200-8.

\bibitem{ahmed2020rgb}
Ahmed A, Jalal A, Kim K.
\newblock RGB-D images for object segmentation, localization and recognition in
  indoor scenes using feature descriptor and Hough voting.
\newblock In: 2020 17th International Bhurban Conference on Applied Sciences
  and Technology (IBCAST). IEEE; 2020. p. 290-5.

\bibitem{zhuang2021semantic}
Zhuang C, Wang Z, Zhao H, Ding H.
\newblock Semantic part segmentation method based 3D object pose estimation
  with RGB-D images for bin-picking.
\newblock Robotics and Computer-Integrated Manufacturing. 2021;68:102086.

\bibitem{wang2020unequal}
Wang J, Xu C, Dai L, Zhang J, Zhong RY.
\newblock An Unequal Learning Approach for 3D Point Cloud Segmentation.
\newblock IEEE Transactions on Industrial Informatics. 2020.

\bibitem{feng2018gvcnn}
Feng Y, Zhang Z, Zhao X, Ji R, Gao Y.
\newblock GVCNN: Group-view convolutional neural networks for 3D shape
  recognition.
\newblock In: Proceedings of the IEEE Conference on Computer Vision and Pattern
  Recognition; 2018. p. 264-72.

\bibitem{mo2019partnet}
Mo K, Zhu S, Chang AX, Yi L, Tripathi S, Guibas LJ, et~al.
\newblock Partnet: A large-scale benchmark for fine-grained and hierarchical
  part-level 3d object understanding.
\newblock In: Proceedings of the IEEE/CVF Conference on Computer Vision and
  Pattern Recognition; 2019. p. 909-18.

\bibitem{yi2017learning}
Yi L, Guibas L, Hertzmann A, Kim VG, Su H, Yumer E.
\newblock Learning hierarchical shape segmentation and labeling from online
  repositories.
\newblock arXiv preprint arXiv:170501661. 2017.

\bibitem{yu2019partnet}
Yu F, Liu K, Zhang Y, Zhu C, Xu K.
\newblock Partnet: A recursive part decomposition network for fine-grained and
  hierarchical shape segmentation.
\newblock In: Proceedings of the IEEE/CVF Conference on Computer Vision and
  Pattern Recognition; 2019. p. 9491-500.

\bibitem{Kalogerakis_2017_CVPR}
Kalogerakis E, Averkiou M, Maji S, Chaudhuri S.
\newblock 3D shape segmentation with projective convolutional networks.
\newblock In: proceedings of the IEEE conference on computer vision and pattern
  recognition; 2017. p. 3779-88.

\bibitem{li2021ctnet}
Li Z, Sun Y, Zhang L, Tang J.
\newblock CTNet: Context-based tandem network for semantic segmentation.
\newblock IEEE Transactions on Pattern Analysis and Machine Intelligence.
  2021;44(12):9904-17.

\bibitem{burgess2019monet}
Burgess CP, Matthey L, Watters N, Kabra R, Higgins I, Botvinick M, et~al.
\newblock Monet: Unsupervised scene decomposition and representation.
\newblock arXiv preprint arXiv:190111390. 2019.

\bibitem{nguyen2019hologan}
Nguyen-Phuoc T, Li C, Theis L, Richardt C, Yang YL.
\newblock Hologan: Unsupervised learning of 3d representations from natural
  images.
\newblock In: Proceedings of the IEEE/CVF International Conference on Computer
  Vision; 2019. p. 7588-97.

\bibitem{10.1007/978-3-319-24574-4_28}
Ronneberger O, Fischer P, Brox T.
\newblock U-Net: Convolutional Networks for Biomedical Image Segmentation.
\newblock In: Medical Image Computing and Computer-Assisted Intervention --
  MICCAI 2015. Cham: Springer International Publishing; 2015. p. 234-41.

\bibitem{2017SegNet}
Badrinarayanan V, Kendall A, Cipolla R.
\newblock SegNet: A Deep Convolutional Encoder-Decoder Architecture for Image
  Segmentation.
\newblock IEEE Transactions on Pattern Analysis Machine Intelligence. 2017:1-1.

\bibitem{2017Pyramid}
Zhao H, Shi J, Qi X, Wang X, Jia J.
\newblock Pyramid scene parsing network.
\newblock In: Proceedings of the IEEE conference on computer vision and pattern
  recognition; 2017. p. 2881-90.

\bibitem{2017Rethinking}
Chen LC, Papandreou G, Schroff F, Adam H.
\newblock Rethinking atrous convolution for semantic image segmentation.
\newblock arXiv preprint arXiv:170605587. 2017.

\bibitem{wold1987principal}
Wold S, Esbensen K, Geladi P.
\newblock Principal component analysis.
\newblock Chemometrics and intelligent laboratory systems. 1987;2(1-3):37-52.

\bibitem{npca}
Papadakis P, Pratikakis I, Perantonis S, Theoharis T.
\newblock Efficient 3D shape matching and retrieval using a concrete radialized
  spherical projection representation.
\newblock Pattern Recognition. 2007;40(9):2437  2452.

\bibitem{wang2019salient}
Wang W, Zhao S, Shen J, Hoi SC, Borji A.
\newblock Salient object detection with pyramid attention and salient edges.
\newblock In: Proceedings of the IEEE Conference on Computer Vision and Pattern
  Recognition; 2019. p. 1448-57.

\bibitem{he2016identity}
He K, Zhang X, Ren S, Sun J.
\newblock Identity mappings in deep residual networks.
\newblock In: European conference on computer vision. Springer; 2016. p.
  630-45.

\bibitem{hamilton2017inductive}
Hamilton W, Ying Z, Leskovec J.
\newblock Inductive representation learning on large graphs.
\newblock In: Advances in neural information processing systems; 2017. p.
  1024-34.

\bibitem{gao20113d}
Gao Y, Dai Q, Wang M, Zhang N.
\newblock 3D model retrieval using weighted bipartite graph matching.
\newblock Signal Processing: Image Communication. 2011;26(1):39-47.

\bibitem{li20193Dor}
Li W, Liu A, Nie W, Song D, Li Y, Wang W, et~al.
\newblock Monocular Image Based 3D Model Retrieval.
\newblock In: 12th Eurographics Workshop on 3D Object Retrieval, 3DORs.
  Eurographics Association; 2019. p. 103-10.

\bibitem{liu2017view}
Liu AA, Nie WZ, Gao Y, Su YT.
\newblock View-based 3-D model retrieval: a benchmark.
\newblock IEEE transactions on cybernetics. 2017;48(3):916-28.

\bibitem{SPH}
Kazhdan M, Funkhouser T, Rusinkiewicz S.
\newblock Rotation invariant spherical harmonic representation of 3 d shape
  descriptors.
\newblock In: Symposium on geometry processing. vol.~6; 2003. p. 156-64.

\bibitem{MVCNN-MultiRes}
Qi CR, Su H, Nie{\ss}ner M, Dai A, Yan M, Guibas LJ.
\newblock Volumetric and multi-view cnns for object classification on 3d data.
\newblock In: Proceedings of the IEEE conference on computer vision and pattern
  recognition; 2016. p. 5648-56.

\bibitem{wang2018dynamic}
Wang Y, Sun Y, Liu Z, Sarma SE, Bronstein MM, Solomon JM.
\newblock Dynamic graph cnn for learning on point clouds.
\newblock arXiv preprint arXiv:180107829. 2018.

\bibitem{chen2003visual}
Chen DY, Tian XP, Shen YT, Ouhyoung M.
\newblock On visual similarity based 3D model retrieval.
\newblock In: Computer graphics forum. vol.~22. Wiley Online Library; 2003. p.
  223-32.

\bibitem{Allen2008VoxNet}
Allen M, Girod L, Newton R, Madden S, Blumstein DT, Estrin D.
\newblock Voxnet: An interactive, rapidly-deployable acoustic monitoring
  platform.
\newblock In: 2008 International Conference on Information Processing in Sensor
  Networks (ipsn 2008). IEEE; 2008. p. 371-82.

\bibitem{VRN}
Brock A, Lim T, Ritchie JM, Weston N.
\newblock Generative and Discriminative Voxel Modeling with Convolutional
  Neural Networks.
\newblock Computer Science. 2016.

\bibitem{yulatent}
Yu Q, Yang C, Fan H, Wei H.
\newblock Latent-MVCNN: 3D shape recognition using multiple views from
  pre-defined or random viewpoints.
\newblock Neural Processing Letters. 2020;52:581-602.

\bibitem{ma2017boosting}
Ma Y, Zheng B, Guo Y, Lei Y, Zhang J.
\newblock Boosting multi-view convolutional neural networks for 3d object
  recognition via view saliency.
\newblock In: Chinese Conference on Image and Graphics Technologies. Springer;
  2017. p. 199-209.

\bibitem{he2019view}
He X, Huang T, Bai S, Bai X.
\newblock View n-gram network for 3D object retrieval.
\newblock In: Proceedings of the IEEE International Conference on Computer
  Vision; 2019. p. 7515-24.

\bibitem{guo2021pct}
Guo MH, Cai JX, Liu ZN, Mu TJ, Martin RR, Hu SM.
\newblock Pct: Point cloud transformer.
\newblock Computational Visual Media. 2021;7:187-99.

\bibitem{gao2021multi}
Gao Z, Zhang Y, Zhang H, Guan W, Feng D, Chen S.
\newblock Multi-level view associative convolution network for view-based 3D
  model retrieval.
\newblock IEEE Transactions on Circuits and Systems for Video Technology.
  2021;32(4):2264-78.

\bibitem{conf/icml/GaninL15}
Ganin Y, Lempitsky VS.
\newblock Unsupervised Domain Adaptation by Backpropagation.
\newblock In: Proceedings of the 32nd International Conference on Machine
  Learning, {ICML} 2015, Lille, France, 6-11 July 2015. vol.~37 of {JMLR}
  Workshop and Conference Proceedings. JMLR.org; 2015. p. 1180-9.

\bibitem{conf/icml/LongZ0J17}
Long M, Zhu H, Wang J, Jordan MI.
\newblock Deep Transfer Learning with Joint Adaptation Networks.
\newblock In: Proceedings of the 34th International Conference on Machine
  Learning, {ICML} 2017, Sydney, NSW, Australia, 6-11 August 2017. vol.~70 of
  Proceedings of Machine Learning Research. {PMLR}; 2017. p. 2208-17.

\bibitem{conf/cvpr/ZhangLO17}
Zhang J, Li W, Ogunbona P.
\newblock Joint Geometrical and Statistical Alignment for Visual Domain
  Adaptation.
\newblock In: 2017 {IEEE} Conference on Computer Vision and Pattern
  Recognition, {CVPR} 2017, Honolulu, HI, USA, July 21-26, 2017. {IEEE}
  Computer Society; 2017. p. 5150-8.

\bibitem{long2013transfer}
Long M, Wang J, Ding G, Sun J, Yu PS.
\newblock Transfer feature learning with joint distribution adaptation.
\newblock In: Proceedings of the IEEE international conference on computer
  vision; 2013. p. 2200-7.

\bibitem{song2023self}
Song D, Zhang CM, Zhao XQ, Wang T, Nie WZ, Li XY, et~al.
\newblock Self-supervised image-based 3d model retrieval.
\newblock ACM Transactions on Multimedia Computing, Communications and
  Applications. 2023;19(2):1-18.

\bibitem{li2023instance}
Li W, Zhang Y, Wang F, Li X, Duan Y, Liu AA.
\newblock Instance-prototype similarity consistency for unsupervised 2D
  image-based 3D model retrieval.
\newblock Information Processing \& Management. 2023;60(4):103372.

\bibitem{wang2018visual}
Wang J, Feng W, Chen Y, Yu H, Huang M, Yu PS.
\newblock Visual domain adaptation with manifold embedded distribution
  alignment.
\newblock In: Proceedings of the 26th ACM international conference on
  Multimedia; 2018. p. 402-10.

\bibitem{li2018pointcnn}
Li Y, Bu R, Sun M, Wu W, Di X, Chen B.
\newblock Pointcnn: Convolution on x-transformed points.
\newblock Advances in neural information processing systems. 2018;31.

\bibitem{kanezaki2018rotationnet}
Kanezaki A, Matsushita Y, Nishida Y.
\newblock Rotationnet: Joint object categorization and pose estimation using
  multiviews from unsupervised viewpoints.
\newblock In: Proceedings of the IEEE Conference on Computer Vision and Pattern
  Recognition; 2018. p. 5010-9.

\bibitem{DBLP:conf/cvpr/HeZRS16}
He K, Zhang X, Ren S, Sun J.
\newblock Deep Residual Learning for Image Recognition.
\newblock In: 2016 {IEEE} Conference on Computer Vision and Pattern
  Recognition, {CVPR} 2016, Las Vegas, NV, USA, June 27-30, 2016. {IEEE}
  Computer Society; 2016. p. 770-8.

\bibitem{DBLP:journals/tcsv/PengC20}
Peng Y, Chi J.
\newblock Unsupervised Cross-Media Retrieval Using Domain Adaptation With Scene
  Graph.
\newblock {IEEE} Trans Circuits Syst Video Technol. 2020;30(11):4368-79.

\bibitem{chen2018text2shape}
Chen K, Choy CB, Savva M, Chang AX, Funkhouser T, Savarese S.
\newblock Text2shape: Generating shapes from natural language by learning joint
  embeddings.
\newblock In: Asian Conference on Computer Vision. Springer; 2018. p. 100-16.

\end{thebibliography}

\end{document}